\documentclass[11pt, a4paper, nonumbering]{llava}
\usepackage{enumitem}
\usepackage[authoryear, compress, round]{natbib}
\usepackage{dblfloatfix}
\usepackage{ulem}
\usepackage{caption}
\usepackage{dramatist}
\usepackage{xspace}
\usepackage{pifont} %
\usepackage{multirow}
\usepackage{tcolorbox}
\usepackage{xltabular}
\usepackage{longtable}
\usepackage{hyperref}
\usepackage{booktabs}
\usepackage{amsmath}

\usepackage{wrapfig}
\usepackage{graphicx}
\usepackage[table]{xcolor} % for cell colors
\usepackage{siunitx}
\usepackage{threeparttable}
\usepackage{lmodern}
\usepackage{amsfonts}
\usepackage{amsmath}
\usepackage{amssymb}
\usepackage{lineno}
\usepackage{multirow}
\usepackage{tikz}
\usepackage[bottom]{footmisc}
\usepackage[utf8]{inputenc}   % 确保UTF-8编码
\usepackage{CJKutf8}
\usepackage{subfigure}
\usepackage{setspace}

\usepackage{makecell}

\usepackage{graphicx}
\usepackage{subcaption}
\usepackage{multicol} %
\usepackage{siunitx}  %
\usepackage{wrapfig}

\usepackage{graphicx}
\usepackage{subfigure}

\definecolor{HeaderBG}{HTML}{F8F9FB}    % Very subtle header background
\definecolor{RowAlt}{HTML}{FDFDFE}      % Extremely light alternate rows
\definecolor{BorderGray}{HTML}{E1E5E9}  % Subtle border color

\usepackage{fontawesome}
\usepackage{xspace}

\newcommand{\eg}{\textit{e.g.}, }
\newcommand{\vs}{\textit{vs.} }
\newcommand{\ie}{\textit{i.e.}, }

\title{\centering  MindGPT-4ov: An Enhanced MLLM via a Multi-Stage Post-Training Paradigm}
\author{
 \textbf{MindGPT-4o Team, Li Auto Inc.}
}
\begin{abstract} 
We present MindGPT-4ov\footnotemark[1], a multimodal large language model (MLLM) that introduces a general post-training paradigm spanning data production, model training, and efficient deployment. It achieves state-of-the-art performance across multiple benchmarks at low cost, effectively enhancing the foundational capabilities of MLLMs and the generalization ability.
Focusing on data construction, supervised fine-tuning strategies, and multimodal reinforcement learning methods, this work proposes three key innovations:
(1) An information density-based data generation scheme, integrated with a dual-dimensional tree-structured label system, enabling automated generation of high-quality cross-domain data. 
(2) A collaborative curriculum supervised fine-tuning approach that balances the injection of domain-specific knowledge with the preservation of general capabilities.
(3) A hybrid reinforcement learning paradigm that enhances reasoning ability while simultaneously addressing multi-objective optimization such as diversity exploration, maintenance of multimodal perception, and response conciseness.
Moreover, we implement a series of infrastructure optimizations, such as 5D parallel training, operator optimization, and inference quantization to enhance training and inference efficiency while reducing the cost of domain adaptation.
Experimental results demonstrate that the MindGPT-4ov model outperforms state-of-the-art models on benchmarks such as MMBench, MMStar, MathVision, and MathVista. In addition, MindGPT-4ov also demonstrates superior user experience in vertical domain tasks, enabling a seamless transition from academic research to industrial deployment. MindGPT-4ov provides a general post-training paradigm applicable to a wide range of MLLMs. The model weights, datasets, and code for the Qwen3-VL-based variants will be recently open-sourced to support the community’s development of MLLMs.

\end{abstract}

\begin{document}

\maketitle

\footnotetext[1]{MindGPT-4o-v(ision) is the visual component of the \href{https://www.lixiang.com/tech/mindgpt}{MindGPT-4o} project led by the foundation model team at Li Auto Inc., which is a multimodal large language model built on MindGPT.}

\section{Introduction}
In recent years, multimodal large language models (MLLMs) have achieved breakthrough progress in vision–language understanding~\citep{gemini2.5,seedv11.5,internvl3, qwen2.5vl, qwen3vl}, demonstrating powerful general perception and reasoning capabilities, which have reached unprecedented performance in tasks such as image understanding, visual question answering, and document analysis. 
A key challenge in current MLLM development is how to effectively transfer general multimodal capabilities to vertical applications while maintaining strong generalization, such as in autonomous driving \citep{tian2024drivevlm, distilling-driving}, embodied intelligence \citep{physvlm, spatialbot, chen2024spatialvlm}, and medical intelligence \citep{healthgpt, lozano2025biomedica}.
To address this challenge, some studies have explored various lines of exploration. For example, SpatialBot \citep{spatialbot} applies MLLMs to embodied intelligence, helping robots perceive their environment and plan tasks; DiMA \citep{distilling-driving} proposes an end-to-end autonomous driving system distilled from a pre-trained MLLM; and HealthGPT \citep{healthgpt} is an MLLM used to support medical treatment.

However, these studies suffer from two problems, resulting in poor practicality and generalization of current models: (1) as MLLMs acquire domain expertise, they often suffer from catastrophic forgetting, degrading their original general understanding capabilities; and (2) they lack a systematic post-training methodology: they either prioritize domain data collection while neglecting stringent data quality and cost controls; focus solely on learning domain-specific capabilities at the expense of maintaining foundational general abilities and real-world user experience; or optimize the training framework without adequately accounting for practical deployment requirements.

By contrast, we propose the MindGPT-4ov framework that offers a systematic, end-to-end methodology that covers the entire pipeline, including: 
\textbf{(1) An information density-driven data synthesis method.} Through automated data acquisition, intelligent annotation assistance, and data quality assessment, we build high-quality training sets with data augmentation and synthesis strategies to alleviate long-tailed data distributions; 
\textbf{(2) A collaborative curriculum supervised fine-tuning method.} This framework integrates domain expertise into the model through coordinated data admission techniques and curriculum learning, which preserves its general capabilities and applies targeted optimizations to address user-experience pain points.
\textbf{(3) A multi-stage hybrid-reward reinforcement learning method.}Through multi-stage reinforcement learning, combined with a carefully designed hybrid reward strategy, we progressively improve both general and vertical domain performance.
\textbf{(4) A 5D parallel training framework.} For Mixture-of-Experts (MoE) architectures \citep{moe_survey}, we introduce a series of training optimizations during the supervised fine-tuning and reinforcement learning phases to significantly improve training efficiency. We also optimize the inference engine to enhance the user experience in high-concurrency and other specific application scenarios.

\begin{figure*}[t]
\centering
\includegraphics[width= 1.0\linewidth]{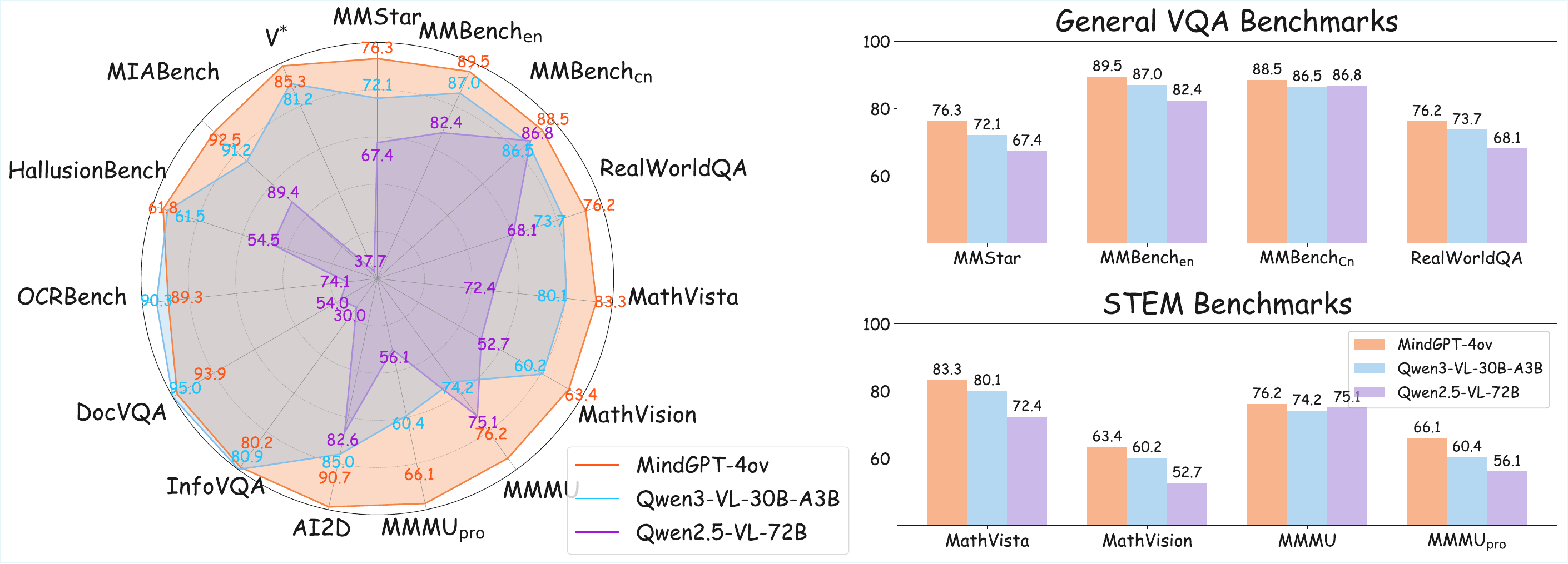}
\caption{Performance of MindGPT-4ov across multiple benchmarks.}
\end{figure*}\label{fig:head_performance}

Ultimately, the model trained with the proposed framework not only excels on vertical applications but also preserves or even enhances general performance, all while providing an enhanced user experience. To be specific, our model performs competitively on general multimodal benchmarks, matching or surpassing state-of-the-art Qwen3-VL-30B-A3B \citep{qwen3vl} across nearly all metrics. In particular, as shown in Fig.~\ref{fig:head_performance}, it exceeds Qwen3-VL-30B-A3B by an average of 3.4\% on general VQA benchmarks and by an average of 3.6\% on math and STEM (Science, Technology, Engineering, Mathematics) benchmarks, respectively. In addition, it also provides a superior real-world user experience, such as  . Finally, the proposed framework supports efficient deployment, characterized by lower CPU/GPU utilization, more efficient cache reuse, faster response times, and improved user experience.

In summary, MindGPT-4ov's end-to-end, systematic design ensures that data preparation, model training, and deployment are all optimized, enabling a seamless transition from academic research to industrial application. Our main contributions include:

\begin{itemize}[leftmargin=*]
    \item \textit{\textbf{A generalizable post-training paradigm.}} We propose MindGPT-4ov, whose training paradigm is applicable to most current MLLMs. Our approach encompasses high-quality data construction, a multi-stage supervised fine-tuning and reinforcement learning scheme, and efficient training and deployment strategies, providing a low-cost, reusable, and scalable technical pathway for the real-world deployment of MLLMs.
    \item \textit{\textbf{Outstanding performance.}} The MindGPT-4ov model achieves superior performance across multiple multimodal general benchmarks compared to Qwen3-VL. Moreover, it can provide a better user experience for the specific domain tasks than Qwen3-VL.
    \item \textit{\textbf{Open-source release.}} We will recently open-source parts of the code and data and models, enabling the community to build on this foundation and extend the models’ capabilities in specific vertical applications.
 \end{itemize}

\section{Data Construction}

\subsection{Challenges}
In the transition from general MLLMs to vertical applications, constructing high-quality training data often encounters several thorny challenges, mainly including:
(1) Producing high-quality data is costly and resource-intensive, overly reliant on manual annotation, and lacks automated pipelines for high-quality data generation.
(2) Training data distributions are often imbalanced: insufficient vertical-domain knowledge data impairs generalization, while a shortage of general multimodal data risks degrading core multimodal capabilities.
(3) Data produced by conventional pipelines often lacks diversity, making it difficult to comprehensively expand multimodal capabilities under limited data budgets.
(4) Vertical-domain data is typically expensive to acquire and relatively scarce, and standard data production methods struggle to unlock the potential value of such scarce data.

\subsection{Solution} \label{sec:solution}

\begin{figure*}[t]
\centering
\includegraphics[width= 1.0\linewidth]{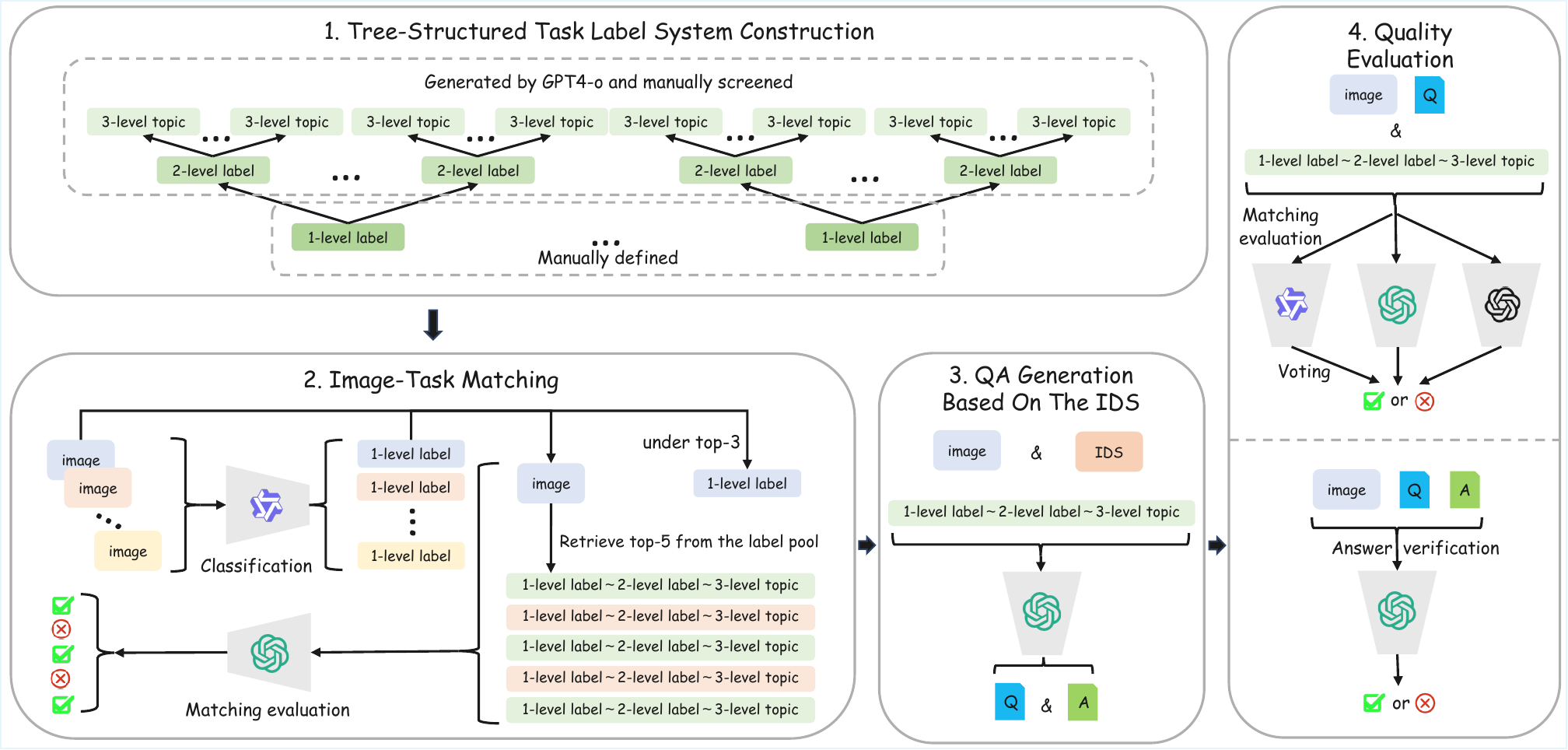}
\caption{An overview of our data pipeline. We begin by constructing a tree‑structured hierarchical label system. We then employ MLLMs to match images with their corresponding task labels. An MLLM is subsequently prompted to generate question–answer pairs based on the images, the task labels, and the information density score (IDS). Lastly, MLLMs are leveraged for matching evaluation and answer verification to ensure data quality.}
\label{fig:data-datapipeline}
\end{figure*}

To address the aforementioned challenges and better support the transition of general-purpose MLLMs to vertical-specific MLLMs, we propose a comprehensive data construction framework, which is presented in Fig.~\ref{fig:data-datapipeline}. The specific innovations are as follows:
We firstly present a dual-dimensional tree-structured label system. This system is designed to balance vertical knowledge coverage and general capability adaptation, which provides the direction for generating data with broad coverage, high diversity, and balanced distribution, while also offering clear guidance for model iteration and evaluation. Then, we develop a data synthesis pipeline based on an information density. Different quantitative systems are designed for static image features, enabling the accurate capture of the potential value of data through weighted fusion. This allows for dynamic adjustment of the quantity of generated QA (question-answer) pairs.

The proposed data production scheme improves upon the coarse-grained indiscriminate resource allocation typical of traditional data synthesis. By generating more high-value data and investing less in low-value data, our method balances resource expenditure while fully tapping the potential of high-quality data. It addresses common challenges in multimodal data production, including low efficiency, poor quality, limited coverage, and imbalance, consistently producing diverse, well-balanced, and highly accurate data, thereby providing an efficient and reliable paradigm for multimodal data generation.

\subsubsection{Dual-dimensional Label System}
To simultaneously support vertical domain knowledge coverage and general capability maintenance, the label system is designed around two dimensions: \textbf{domain} and \textbf{ability}.

\begin{itemize}

\item{Domain dimension} covers vertical domain knowledge across automotive, healthcare, education, industry, daily life, and others, anchoring each datum to concrete application scenarios to ensure the synthesized data aligns with target domains. 

\item{Ability dimension} focuses on the general abilities of MLLMs, primarily encompassing image-related dimensions such as OCR, counting, spatial perception, logical reasoning tasks, and others.

\end{itemize}

The dual-dimensional design offers the advantage of breaking through the limitations of traditional single-dimensional label systems. It enables each synthesized data sample to be associated with specific vertical domain knowledge while simultaneously addressing explicit model ability training requirements. This establishes a foundation for subsequent multimodal training data synthesis. The comprehensive and diverse domain labels help guide the data production in synthesizing domain knowledge tailored to various verticals. Decoupling knowledge from ability also allows the training process to better understand data distribution, facilitating adjustments in data allocation during training and ensuring both domain-specific adaptation and the retention of general multimodal capabilities.

\begin{figure*}[t]
\centering
\includegraphics[width= 1.0\linewidth]{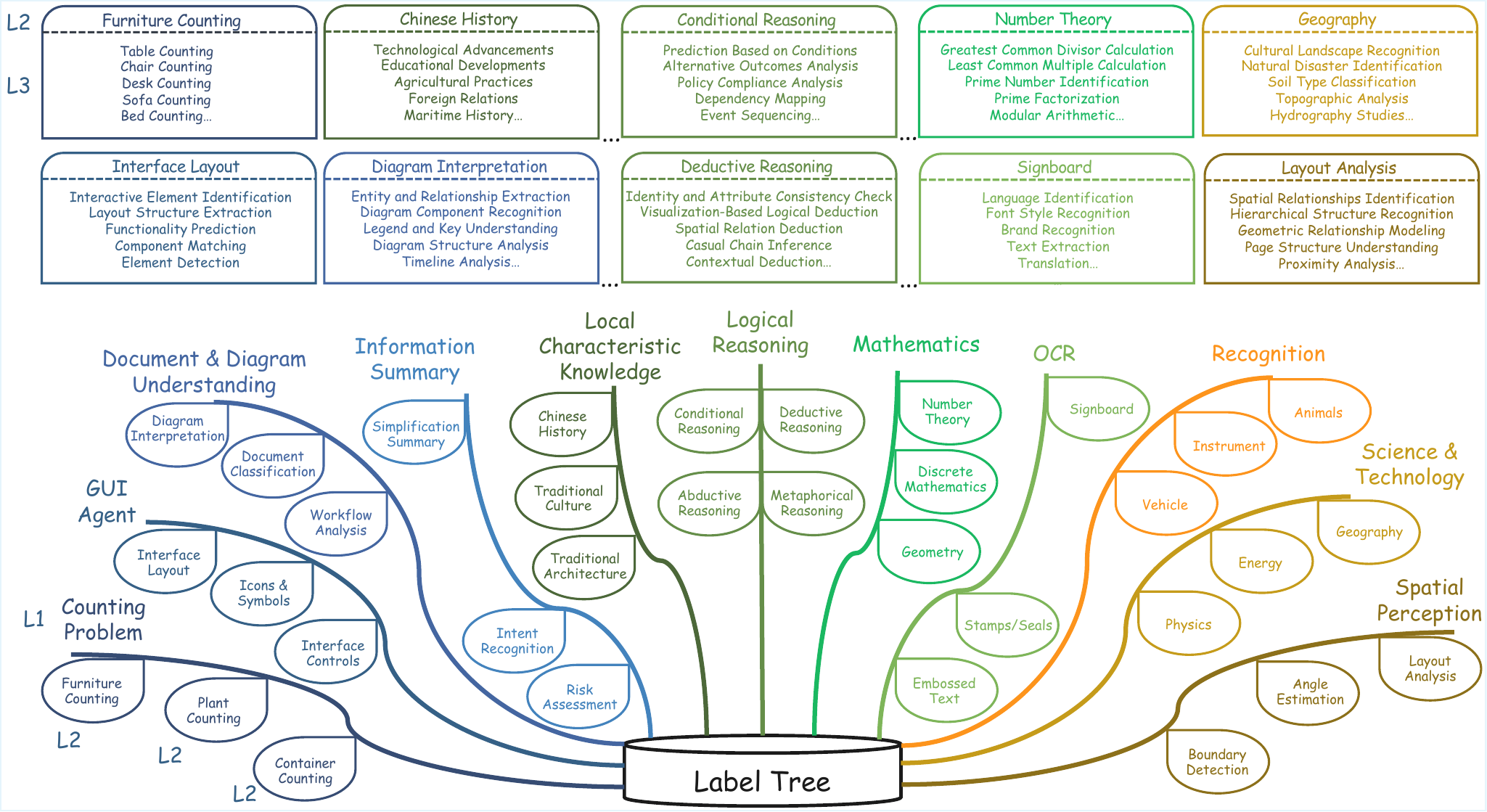}
\caption{The tree-structured label system. In the label tree, each branch denotes a first‑level label, while the leaves represent a selection of corresponding second‑level labels. The box diagrams above the tree illustrate representative third‑level topics encompassed by certain second‑level labels.}
\label{fig:label_tree}
\end{figure*}

\subsubsection{Topic Tree}
Inspired by \citep{chentaskgalaxy}, to address the limited coverage and insufficient richness of manually constructed labels of training data, as shown in Fig.~\ref{fig:label_tree}, we construct a topic tree through an expert-defined taxonomy and model-based label expansion. Based on the dual dimensions, we construct corresponding two-level labels and ensure their reliability through human curation, forming a foundational knowledge and ability label system. Building upon the two-level labels, we leverage MLLMs to construct topics that combine domain knowledge and abilities, aiming to comprehensively cover the specific applications of MLLMs' abilities in vertical domains. Ultimately, we arrive at a tree structure for both domain and ability dimensions: first-level ($L_1$) $\sim$ second-level ($L_2$) labels and third-level topic ($L_3$). The process specifically consists of the following three steps: 

\textbf{(1) Expert definition.} Experts manually define the $L_1$ label for both domain and ability dimensions to ensure reliability and coverage.

\textbf{(2) Label expansion.} We carefully design prompts and leverage GPT-4o \citep{gpt-4o} to expand $L_2$labels based on manually defined $L_1$ labels, aiming to maximize coverage while maintaining decoupled label definitions.

\textbf{(3) Topic generation.} Based on $L_1$ and $L_2$ labels, we use GPT-4o to generate more fine-grained $L_3$ topics, ensuring that each topic reflects the concrete application domains of specific general capabilities, \eg \textit{Counting Problem $\sim$ Symbol/Sign Counting $\sim$ Traffic Signal Counting}.

Leveraging the semantic comprehension and generative capabilities of MLLMs, we obtain 38,455 $L_3$ topics based on predefined $L_1$ and expanded $L_2$ labels. The construction of this extensive topic tree has significantly mitigated the limitations of manually defined labels, such as constrained coverage and inadequate richness. Meanwhile, the topics fuse ability and domain knowledge into practical applications, highlighting the close connection between multimodal capabilities and vertical scenarios and providing support for synthesizing rich and diverse data.

\subsubsection{Topic Matching} \label{sec:topic_matching}
The goal of topic matching is to accurately associate multimodal data with relevant topics, ensuring the precision and effectiveness of synthetic data. This process leverages the inherent content strengths of the data for annotation to prevent scenarios where high-value data results in low-quality QA pairs or misaligned content. Moreover, due to the vast scale and rich content of the topic tree we construct, the matching process ensures from the source that the generated data achieves broad coverage and high diversity. Specifically, we employ a two-stage matching process of coarse-grained matching followed by fine-grained matching, leveraging the powerful comprehension and multimodal retrieval capabilities of MLLMs, and assign each data corresponding topics. The specific approach is as follows:

\textbf{Coarse-grained matching.} We employ high-performance MLLMs, such as Qwen3VL-235B-A22B \citep{qwen3vl}, to perform relevance ranking of $L_1$ labels, achieving preliminary matching between image content and labels. The procedure is as follows: Provide the image along with the full set of $L_1$ labels, and include an \textit{Others} category as a boundary for relevance ranking to prevent ambiguously borderline data from being forcibly categorized. Then rank labels by their relevance to the image content and select the Top-3 $L_1$ labels that fall above the \textit{Others} category.

To validate the accuracy of coarse-grained matching, we construct a small evaluation set in which each image contains 2 or 3 manually annotated $L_1$ labels. If the Top-3 $L_1$ labels resulting from the matching process overlap with the annotated labels, the matching is considered valid. Evaluation results show that the coarse-grained matching achieves an accuracy of 89.7\%, which provides a reliable foundation for the subsequent fine-grained matching.

\textbf{Fine-grained matching.} Within the scope of the selected $L_1$ labels, we use a multimodal retrieval model (\eg Ops-MM-embedding-v1-7B \citep{Ops-MM-embedding-v1-7B}) to compute cross-modal vector similarity between the $L_1 \sim L_2 \sim L_3$ topics and images and retrieve the Top-5 related $L_1 \sim L_2 \sim L_3$ topics as candidate topics for each image.

\subsubsection{Information-density-driven Data Synthesis}
Acquiring informative vertical domain data is highly challenging, as it is often difficult to accumulate in large quantities. To fully leverage the potential of scarce data, we propose a data synthesis solution driven by \textbf{Information Density Score (IDS)}. By quantifying the potential value of each data point, we dynamically adjust the number of generated QA pairs for them, producing more for high-value data and fewer for low-value data. This method proves more resource-efficient than traditional coarse-grained synthesis approaches, which often allocate resources uniformly. Our method prioritizes high-value data while maintaining cost-effectiveness. Furthermore, to ensure high quality of produced data, we implement a multi-model voting mechanism for post-synthesis quality assessment. This significantly reduces the need for manual intervention and provides a foundation for cost-effective synthetic vertical-domain data.

\textbf{Data synthesis based on IDS.}
The IDS assesses image data across four dimensions: subject diversity (the granularity and richness of objects and scene elements in the image), spatial relationship of scenes (the complexity of relative positions and hierarchical relationships among objects), OCR text dimension (the richness of textual information in the image), and world knowledge relevance (the density of entity information in the image that requires external knowledge to answer).

We use MLLMs to score candidate images for QA synthesis based on these four dimensions. The scores are then weighted and fused according to the contribution of each dimension. During weight allocation, the contribution of each dimension to multimodal capability is considered in combination with the application scenarios, ultimately resulting in a unified IDS.

The candidate topics obtained through topic matching determine the direction of QA synthesis, while the IDS determines the upper limit of the number of QA pairs to be generated per image. Images with higher IDS yield more QA pairs, while those with lower IDS yield fewer. QA synthesis is guided by carefully designed prompts and generated based on the candidate topics. Special processing is applied to data with existing annotations to ensure that the original annotation information is not lost.

\textbf{Quality evaluation based on judge voting.} To better align images, topics, and questions, we employ three high-performance MLLMs, Qwen3VL-235B-A22B, GPT-4o, and GPT-5 \citep{gpt5}, as judges to evaluate whether the topics and questions match the given images. Specifically, each of these three MLLMs provides a score of 0 or 1 to every (image, topic, question) triplet based on its consistency and rationality. Only samples that receive a score of 1 from at least two models are retained.

This voting mechanism effectively reduces evaluation errors and ensures the accuracy of (image, topic, and question) matching. Additionally, the quality of responses is equally critical. Therefore, we further leverage high-performance MLLMs to validate the generated responses. We instruct the MLLMs to evaluate whether the generated responses are rational and consistent with the images based on five dimensions: accuracy, completeness, authenticity, semantic relevance, and detail plausibility. This process allows filtering and retaining data with high-quality responses.
\section{Collaborative Curriculum Supervised Fine-Tuning}
Traditional supervised fine-tuning (SFT) often relies on randomly mixing large-scale data, which leads to three major problems:
(1) Learning is imbalanced between domain knowledge and multimodal capabilities: overemphasis on cross-domain knowledge acquisition can degrade multimodal abilities, while forced reasoning with insufficient domain knowledge leads to hallucinations.
(2) It fails to target model weaknesses: coordination between data production and training is inefficient, resources are often spent on already-mastered capabilities, and training costs are high.
(3) It is not optimized for user experience in real vertical scenarios, leading to poor user experience.

To address the above issues, we propose a curriculum-learning SFT paradigm that collaborates with data admission. First, we design a flexible data admission scheme applied throughout the training phases, effectively improving the synergy between data production and model training while reducing overall training costs. Next, we introduce a three-stage progressive SFT strategy: we begin with cross-domain knowledge learning and subsequently repair degraded multimodal capabilities. Finally, to mitigate suboptimal user experience in real-world usage, we perform targeted optimizations and improvements. We focus on the following training objectives:
(1) Improve vertical knowledge learning, and bridge the model's knowledge gaps in domains.
(2) Balance perception and reasoning capabilities learning, and mitigate capability decay caused by cross-domain data distribution differences or insufficient data quality.
(3) Improve user experience through customization tailored to specific domains.

\subsection{Data Preparation} \label{sec:sft_data_pre}
To support the three-stage curriculum learning process, we split the whole training data into three subsets: \textbf{SFT-Knowledge}, \textbf{SFT-Ability}, and \textbf{SFT-Preference}. These subsets have different focuses: SFT-Knowledge targets the acquisition of new knowledge in vertical domains; the images of the SFT-Knowledge set are sourced from real-world usage scenarios, whose data annotations are obtained using the data synthesis method described in Sec.~\ref{sec:solution}.
SFT-Ability focuses on enhancing general capabilities. The images of the SFT-Ability set are sampled from open-source datasets. To collect as much of such data as possible, we sample data from a large-scale repository of images from open-source datasets, including Laion \citep{laion}, SA-1B \citep{sam}, and others. We use the data synthesis pipeline described in Sec.~\ref{sec:solution} to annotate the collected images. SFT-Preference is responsible for improving the user experience. The data of the SFT-Preference set are sampled from specific domain or open-source datasets. 

\subsection{Training Recipe}

\subsubsection{Overview}
This chapter focuses on the data admission pipeline for providing high-quality data and the three-stage curriculum-based model training process. These two components work in synergy to collectively optimize the performance of the SFT model.

\begin{figure*}[t]
\centering
\includegraphics[width= 1.0\linewidth]{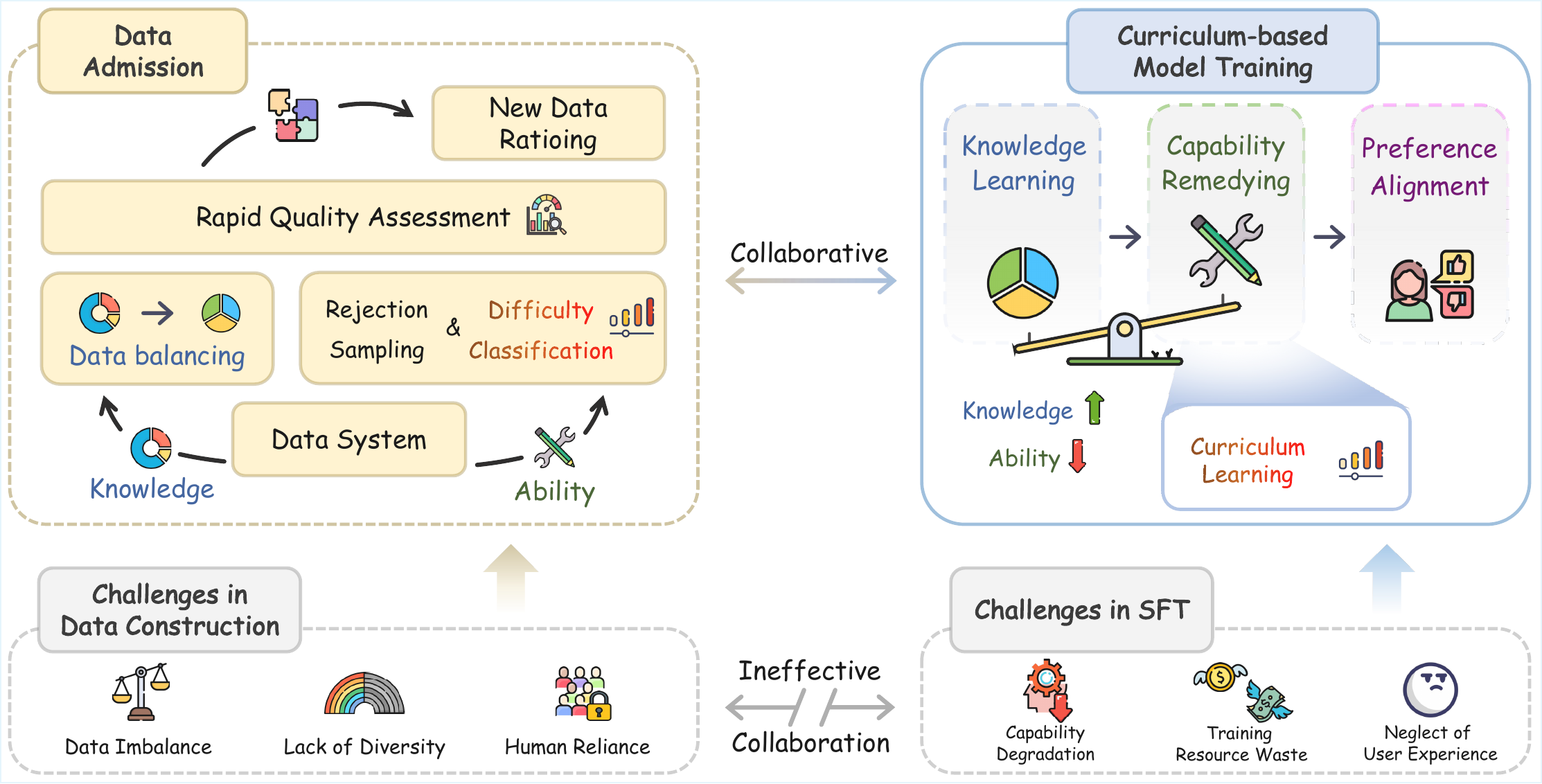}
\caption{Overview of collaborative curriculum supervised fine-tuning (SFT). The proposed SFT training paradigm has the following advantages: on the data side, it automates dataset construction and produces balanced, diverse data; on the training side, it maintains a balanced development of knowledge and capabilities, remedies weak abilities, and safeguards the user experience. Meanwhile, the data and training sides collaborate to substantially improve the efficiency and stability of SFT training.}
\label{fig:sft-framework}
\end{figure*}

As shown in Fig.~\ref{fig:sft-framework}, the data admission pipeline begins by performing data balancing and rejection sampling on the collected data. Data balancing is designed to achieve an even distribution of data across different capability dimensions, while rejection sampling is employed to grade the difficulty of the data. Subsequently, both processed data components undergo a data quality evaluation. Furthermore, we have established an engineered validation mechanism for newly introduced data, enabling efficient integration of multi-source new data into the training set to effectively expand its scale.

The three-stage curriculum-based SFT training process consists of three parts: knowledge learning, capability remedying, and preference alignment. Knowledge learning focuses on injecting vertical domain knowledge to equip the model with the ability to solve domain-specific problems; capability remedying aims to recover the model's performance on general tasks; and preference alignment is used to enhance the model's performance in real-world application scenarios and improve the user experience.

\subsubsection{Data Admission}
Fine-tuning MLLMs typically require combining diverse data sources from both general and vertical corpora, making it difficult to find an effective data-mix ratio that achieves balance. At the same time, vertical data across domains are heterogeneous and vary widely in quality, so efficiently screening data quality is one of the challenges in training iterations. In practice, model training and data production often proceed in parallel; both sides need close, efficient collaboration and a rapid feedback loop to support fast model iterations. The training data admission process can quickly identify data quality, realize an optimally balanced data mix, and provide efficient feedback to the data production side, greatly reducing the frequency of data validation experiments and accelerating overall model iteration.

\textbf{Data balancing.} 
SFT data is sourced from both open-source and collected real vertical datasets. We apply the label matching process mentioned in Sec.~\ref{sec:topic_matching} to the open-source data and manage it uniformly with vertical data. When vertical data is severely imbalanced or scarce, our two-level labels and specific topic enable rapid sampling in data-scarce dimensions to achieve balanced training data. To mitigate noise from imprecise label matching, we apply a latent class-level clustering algorithm to the balanced training data. The distribution of data volumes across different clusters helps evaluate whether optimal balance has been achieved.

\textbf{Rejection sampling.} During the continuous training iterations of the model, we also employ a rejection sampling process to further refine the data selection. As vertical domain knowledge is successfully incorporated, we use rejection sampling to evaluate the training data, categorizing it into three levels: proficiently mastered, partially mastered, and not mastered. In subsequent model iterations, we increase resampling for partially mastered and unmastered data while downsampling proficiently mastered data. This allows the ongoing training to continually focus on capabilities the model has not yet acquired, promoting faster model improvement and more efficient use of training resources.

\textbf{Rapid quality assessment.} The balanced training data is partitioned according to its intended use. We perform rapid SFT training on the base model using small batches of data, focusing on the model’s performance after training. Data admission is determined based on two criteria: whether general performance metrics are satisfactorily maintained and whether vertical capabilities are improved. Typically, we conduct a case-by-case analysis of the results and provide feedback to the data production side to refine subsequent production and sampling strategies.

\textbf{New data ratioing.} As both open-source and vertical data continue to grow, we maintain stable and efficient iteration of training data through a dynamic data ratio allocation process. To accelerate validation, we sample small batches of balanced data from a stable data version as the base, then blend it with newly added data to form updated ratios. The training effect is promptly validated through rapid training with early stopping, and by analyzing metrics and failure cases, we quickly determine the optimal ratio for the next training cycle. Eventually, the newly incorporated data is fully integrated into the new stable data version.

\subsubsection{Curriculum-based Model Training}
The overall SFT training process is structured around two key dimensions.
The first dimension divides the training into three stages based on \textbf{cognitive progression}, balancing the trade-off between vertical domain knowledge and general capabilities. This curriculum-based approach mirrors human cognitive development: it begins with understanding the world (knowledge acquisition), then moves to strengthening weak areas (capability building), and finally advances to integrating diverse skills and generalizing knowledge (flexible adaptation to vertical scenarios).
The second dimension involves \textbf{difficulty gradation} within each stage. Throughout the training, we refine the data into three difficulty levels via rejection sampling, enabling steady improvement in the model’s capabilities, guiding its evolution from foundational to advanced proficiency while eliminating weakness.
Based on the principles outlined above, we divide the SFT training process into the following three steps:

\textbf{Stage-1: Cross-domain knowledge learning.} 
Two key factors drive our decision to prioritize cross-domain knowledge learning in the SFT phase. First, the cognitive sequence should progress from perception to reasoning. Therefore, during early model training, we focus on enhancing the model's vertical domain knowledge to prevent it from generating significant hallucinations when performing reasoning in new domains without adequate domain-specific knowledge. Second, vertical knowledge data often comes from heterogeneous sources with varying and uncontrollable quality, which can adversely affect the model's capabilities and lead to severe degradation in general multimodal benchmark performance. To balance the trade-off between vertical domain knowledge and general capabilities, we first conduct cross-domain knowledge learning, followed by a dedicated phase to restore degraded general abilities. Additionally, the training data for this stage is designated as SFT-Knowledge, which focuses extensively on vertical knowledge. Its specific production methodology is detailed in Sec.~\ref{sec:sft_data_pre}.

\textbf{Stage-2: Capability remedying.}
To address the general capability degradation issue after cross-domain knowledge learning, this stage utilizes the SFT-Ability dataset, which is specifically tailored for models that have undergone cross-domain knowledge acquisition. The detailed production method is introduced in Sec.~\ref{sec:sft_data_pre}. Prior to training, we also employ an admission strategy to sample from the SFT-Ability data and supplement underperforming areas, creating custom blending ratios based on the severity of deficiencies in different capabilities. For example, if the model from the first stage exhibits a decline in accuracy on STEM benchmarks, we can supplement it with "Science and Technology" and "Math" capability data produced through the data production pipeline proposed in Sec.~\ref{sec:solution}.

\textbf{Stage-3: Preference alignment.}
The objective of this training phase is to align with human preference in vertical scenarios through training with extremely high-quality data, while simultaneously consolidating the model’s general multimodal capabilities and establishing a foundation for subsequent reinforcement learning. During this stage, a finely constructed SFT-Preference data subset is employed. This subset consists primarily of the following components:

\begin{itemize}

\item \textbf{Single-modal spurious-correlation data}. 
Current multimodal model architectures, which are predominantly LLM-centric, often over-rely on linguistic capabilities, leading to unimodal spurious correlations and consequently generating significant response hallucinations in practical applications \citep{leng2024curse, li2025devil}. To mitigate this problem, we enhance training with specifically constructed data, such as image-free question-answer pairs, image-text irrelevant question-answer pairs, and multi-turn irrelevant dialogue data, thereby strengthening both multimodal and linguistic capabilities.

\item \textbf{Instruction-preference data}. For different vertical domain scenarios, we include preference-tailored data to shape distinct response format styles (\eg structured markdown and human-like styles).

\item \textbf{Long context data}. By increasing the training token limit so the model can ingest and produce longer sequences, we enhance logical reasoning and lay the foundation for subsequent reinforcement learning.

\item \textbf{High-difficulty supplementary data}. We employ rejection sampling to extract data that the model completely fails to handle from the reinforcement learning phase (described in Sec.~\ref{sec:rl}), allowing the model to learn it in advance through SFT to lay the groundwork for subsequent reinforcement learning.
\end{itemize}

\section{Hybrid Reinforcement Learning} \label{sec:rl} 

After the SFT stage, the model acquires initial domain knowledge and achieves a balanced level of multimodal capability. Considering the demonstrated effectiveness and efficiency of the recent post‑training paradigm based on RLVR (Reinforcement Learning with Verifiable Rewards) \cite{cao2024survey,zhang2025survey,guo2025deepseek,shao2024deepseekmath,Glm-4.1v,seedv11.5}, we further employ RLVR to enhance the model’s reasoning abilities and generalization performance on complex multimodal tasks. However, the conventional RLVR training paradigm faces several bottlenecks when applied to multimodal tasks:

\begin{itemize}
    
\item  Exploration-exploitation imbalance and reduced generalization.
Existing RLVR reward designs typically optimize the Pass@1 objective. While this raises average performance on standard benchmarks, it biases the model toward imitating previously high-reward behaviors and avoiding low-reward actions. Such imitation narrows the model’s effective inference ability frontier ~\citep{yue2025does,chen2025pass}: training does not introduce genuinely new reasoning capabilities and instead suppresses exploratory behavior, leading to the degraded overall generalization.

\item  Accuracy-diversity trade-off and cognitive rigidification.
Verifiable and correctness-focused rewards tend to prioritize accuracy at the expense of diversity ~\citep{li2025jointly,song2025outcome}. To maximize measured correctness, training progressively sharpens the output distribution, causing the model’s outputs to converge and the information content to drop substantially. This diversity collapse severely limits the utility of MLLMs in exploratory applications (\eg creative writing, scientific discovery and open-ended problem solving), and reduces the model’s ability to generalize on domain-specific tasks.

\item  Advantage-signal collapse during Reinforcement Learning (RL) training.
When within-group response rewards become uniform (all-correct or all-incorrect), the advantage signal vanishes and gradient information becomes ineffective ~\citep{yu2025dapo,Glm-4.1v}. This reduces the proportion of informative queries, lowers training efficiency, and impedes the model’s ability to explore deep, multi-step reasoning.

\end{itemize}

To alleviate the above issues, we propose a multimodal reinforcement-learning paradigm whose core components are:

\begin{itemize}

\item  Targeted data collection and cleaning combined with rejection sampling, yielding a high-quality RL dataset that supplies both general and domain knowledge, and alleviates training inefficiency caused by advantage collapse.

\item  A two-stage online-offline hybrid RL scheme that balances text-only and multimodal competencies, progressively improving both general-purpose and domain-specific metrics while mitigating visual neglect.

\item A multi-dimensional hybrid reward design that simultaneously promotes improvements in model’s overall performance, encourages exploratory diversity and concise replies.
\end{itemize}

\subsection{Data Preparation} 
\begin{figure}[ht]
\centering
\includegraphics[width= 0.9 \linewidth]{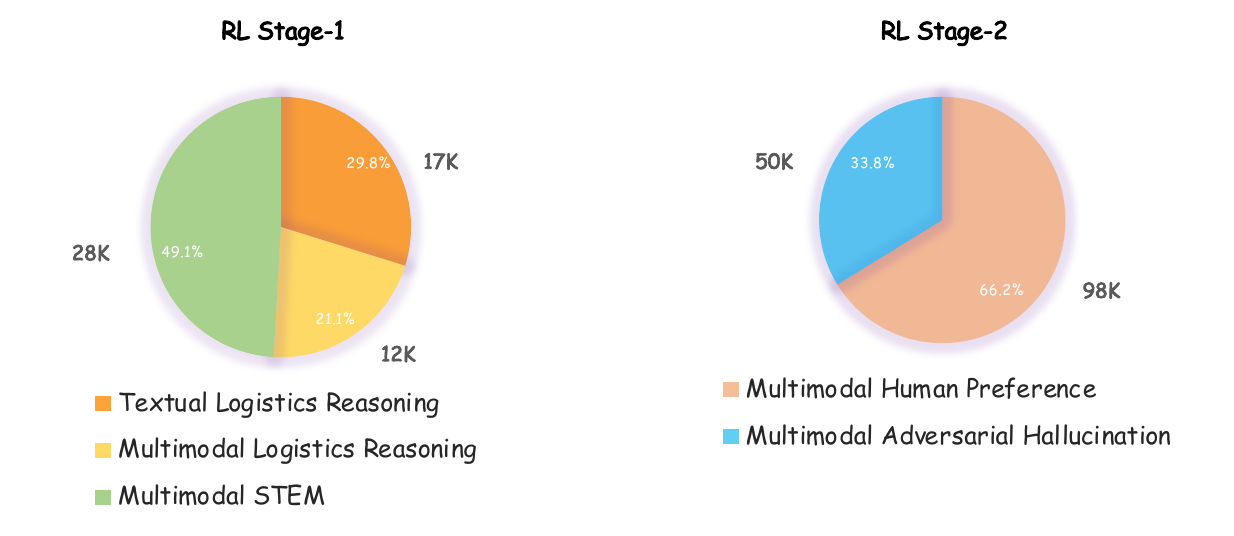}
\caption{\label{fig:4-1} The distribution of RL training data.}
\end{figure}

This section details the RL data construction pipeline. During the first online RL stage, we assemble multimodal logical-reasoning and multidisciplinary STEM corpora to enhance the model’s capacity for complex reasoning. During the second offline RL stage, we prepare targeted human preference data to improve domain performance, and we further construct a set of adversarial hallucination examples to suppress hallucination behaviors. In total, we construct 205K reinforcement-learning examples; their distribution is shown in Fig.~\ref{fig:4-1}.

\subsubsection{Online RL Data}
\textbf{Logical-reasoning}. Logical-reasoning examples are critical for RL-driven improvement ~\citep{seedv11.5,Glm-4.1v,Qwen-VL,qwen2.5vl,wang2024qwen2,structvrm,kimiteam2025kimivltechnicalreport,internvl3,wang2025internvl3_5}. Accordingly, we perform large-scale collection from multiple open-source datasets and retain only the highest-quality subsets after per-dataset quality validation. Each source dataset is validated independently to verify its suitability; passing subsets underwent manual screening to remove manifestly incorrect instances prior to inclusion in the RL data pool. To mitigate reward hacking, all multiple-choice items are converted to open cloze (fill-in-the-blank) format ~\citep{structvrm}. To prevent forgetting of SFT-acquired knowledge, we randomly sample a portion of logical reasoning problems from the preceding SFT training set and mix them into the RL training data. Finally, to avoid degradation of the pure-text LLM capabilities embedded in the MLLM, we include a dedicated set of text-only logical reasoning problems to maintain the LLM component’s reasoning competence.

\textbf{STEM}. The STEM corpus is constructed to broaden the model’s multidisciplinary competence while further improving reasoning \cite{structvrm,Glm-4.1v,nvidia-scaling}. We collect document sources from open web repositories and feed the raw documents into the open-source parser MinerU \cite{niu2025mineru2} to extract candidate (image, question, answer) triples. During extraction, we exclude documents lacking images and retain only question instances that require image evidence for correct resolution. Because MinerU occasionally produces extraction artifacts (OCR errors, layout-induced sentence reorderings, and other hallucinations), we apply a large MLLM as a second-pass filter to cleanse and remove samples exhibiting such defects.

To improve the efficiency and stability of online RL and to mitigate advantage-signal collapse, we apply rejection sampling to exclude trivially all-correct or all-incorrect training examples \cite{structvrm,yu2025dapo}. Additionally, for precise answer parsing and accurate reward computation, we append the following instruction to the end of every training sample:\textit{ The final answer MUST BE put in  $\backslash \backslash boxed\{ \}$}. This constraint forces the model to output the final answer at a designated location, enabling reliable automatic scoring during RL.

\subsubsection{Offline RL Data}

\textbf{Adversarial hallucination data.} To mitigate hallucinations stemming from unimodal knowledge leakage or memorization (\ie the model confidently answering without visual evidence) \cite{hallucinationsurvey,hallucinationsurvey2,unimodalspurious}, we construct a dedicated adversarial hallucination dataset. The construction pipeline proceeds as follows. First, we create image-ablated variants of existing image–text QA pairs by removing the image input and retaining only the question text; the model is then required to regenerate the answer under the image-absent condition. These image-ablated examples explicitly penalize and correct unsupported assertions by training the model not to fabricate visual facts when visual evidence is unavailable. Second, we automatically screen the original multimodal QA samples: if removal of the image still produces a response that is highly consistent with the original ground truth, we mark the sample as exhibiting LLM knowledge leakage (\ie the answer is language-driven rather than visually grounded). Such samples are excluded from the multimodal RL pool to prevent the RL phase from amplifying hallucination behaviors. The resulting adversarial dataset encourages reliance on visual evidence during reasoning and reduces false answers in downstream deployments.

\textbf{Human preference data.} Human preference-based RL data are designed to optimize domain-specific experience by teaching the model to generate responses that better match human preference in real-world scenarios. This dataset spans multiple response styles (\eg concise and formal) and contains explicit annotations for attributes such as response length, fluency, politeness, and domain compliance. It also includes corrective preference annotations addressing common failure modes (\eg overly long replies, stilted phrasing, and redundant information).

\subsection{Training Recipe}
\subsubsection{Overview}
\begin{figure}[t]
\centering
\includegraphics[width= 0.99\linewidth]{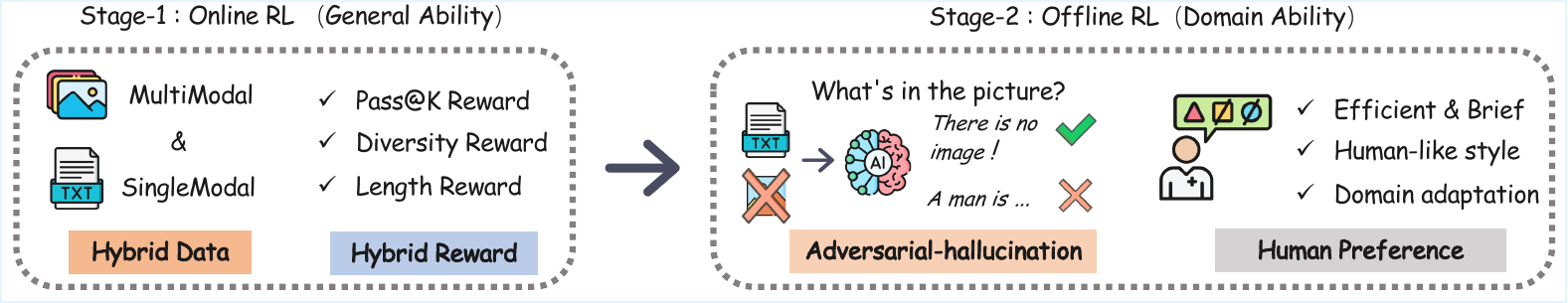}
\caption{\label{fig:4-2} The overview of two-stage hybrid RL training.}
\end{figure}

In this section, we present a multi-stage reinforcement-learning protocol designed to simultaneously strengthen the model’s general-purpose reasoning and domain-specific capabilities. The overall pipeline is illustrated in Fig.~\ref{fig:4-2} .

\textbf{Stage-1: Online RL for generality and diversity.}
We perform GSPO-based RLVR \cite{gspo} and extend the conventional RL reward scheme to explicitly promote exploration and output diversity. Concretely, we firstly incorporate a Pass@k-based reward \cite{chen2025pass} to incentivize exploration during the early phase of training, and then introduce a diversity reward \cite{li2025jointly} in the mid-to-late phase to encourage varied model outputs. This schedule deeply stimulates the model’s multimodal reasoning while preserving and expanding its exploratory behavior and generalization.

\textbf{Stage-2: Offline RL for domain ability.}
We perform DPO (Direct Preference Optimization) \cite{dpo} using a carefully curated hybrid preference corpus. This offline preference-alignment stage reduces hallucinations and aligns outputs with application scenario preferences while maintaining strong multimodal ability, output diversity, and overall user experience.

\subsubsection{Hybrid Reward}

To alleviate the limitations of reward design in conventional RLVR, we develop a hybrid reward strategy that enables the model to strike a balance among exploratory diversity, preservation of multimodal perception capabilities, and reduction of response redundancy when serving domain‑specific applications. At the same time, this approach lowers inference costs and significantly improves real‑world user experience. In this section, we provide a detailed introduction to the proposed reward system.

\textbf{Pass@1 Reward.} For RLVR tasks the most commonly adopted metric is the Pass@1 reward \cite{guo2025deepseek,shao2024deepseekmath}, which compares the model prediction with the ground-truth answer and grants a unit reward only for an exact match. Formally, letting $\hat{y}$ denote the model output and $y$ the ground truth, we define:
\begin{equation}
r_{\text{Pass@1}} =
\begin{cases}
1, & \text{if } \  \hat{y} = y  \\
0, & \text{otherwise.}
\end{cases}
\end{equation}

\textbf{Pass@k Reward.} However, when deployed on complex real-world tasks, a Pass@1-only strategy for multimodal RL exhibits several fundamental limitations \cite{chen2025pass,yue2025does}. A monotonic, correctness-only reward signal tends to bias optimization toward “accuracy at the expense of diversity.” Under this single-minded pressure the model’s reasoning patterns gradually rigidify: to maximize rewards it converges on a narrow set of safe, highly ranked responses, producing homogenized outputs (also known as \textit{diversity collapse}). This collapse undermines user experience (users expect creative, varied answers rather than repetitive replies) and even may erode the model’s core multimodal capabilities. 

To alleviate the above issue, we employ the Pass@k reward scheme \cite{chen2025pass}. For a given query x, let the policy $\pi_\theta$ generate k responses; denote the $i$-th sampled response by $\hat{y}_i$, which receives a verifier-provided reward $R_i$. Pass@k is defined as the expected maximum reward among the $k$ samples:
\begin{equation}
\mathrm{Pass@}k = 
\mathbb{E}_{(x,y) \sim \mathcal{D}, \{\hat{y}_i\}_{i=1}^k \sim \pi_\theta(\cdot \mid x)} \left[ \max\left(R_1, R_2, \dots, R_k\right) \right].
\end{equation}

Compared with Pass@1, Pass@k permits the policy to produce multiple (possibly incorrect) candidates and thus explicitly rewards generation of diverse responses that cover a larger portion of the solution space, improving exploration. 
Concretely, when computing the advantage for each sampled response, we follow the grouping-based derivation used in the original Pass@k analysis \cite{chen2025pass}: We partition the $N_{rollout}$ generated responses into groups of size $k$ and compute each group’s mean reward and standard deviation, which are then used to normalize per-sample advantages. The number of groups is calculated as ${N_{total}^{group}} = \binom{N_{\text{rollout}}}{k}$, and the number of negative groups can be calculated as ${N_{neg}^{group}} = \binom{N_{\text{neg}}}{k}$. Then, we obtain the number of the positive groups ${N_{pos}^{group}} = \binom{N_{\text{rollout}}}{k} - \binom{N_{\text{neg}}}{k}$. If we set positive reward as 1 and negative as 0, we can obtain the average rewards of the group $\bar{R}^{group}$:
\begin{equation}
\overline{R}^{\text{group}} = 1 - \frac{\binom{N_{\text{neg}}}{k}}{\binom{N_{\text{rollout}}}{k}}.
\end{equation}
The standard variance $\sigma^{\text{group}}$ can be calculated as follows:
\begin{equation}
\sigma^{\text{group}} = \sqrt{\overline{R}^{\text{group}} \times \left( 1 - \overline{R}^{\text{group}} \right)}.
\end{equation}
A positive response appears only in groups that each contain at least one correct answer. A negative response can occur either in such mixed (positive) groups or in purely incorrect (negative) groups. Consequently, we deduce the advantage of the positive group $\hat{A}_{\text{pos}} $ via the standard variance $\sigma^{\text{group}}$, which can be calculated as follows:
\begin{equation}\label{eq:5}
\hat{A}_{\text{pos}} = \left(1 - \overline{R}^{\text{group}}\right) \times \left( \sigma^{\text{group}} \right)^{-1},
\end{equation}
and the negative group $\hat{A}_{\text{neg}} $ is calculated as follows:
\begin{equation}\label{eq:6}
\hat{A}_{\text{pos}} = \left(1 - \overline{R}^{\text{group}}\right) \times \left( \sigma^{\text{group}} \right)^{-1}.
\end{equation}
\textbf{Diversity Reward.} To jointly optimize quality and diversity during RL, we introduce a diversity reward \cite{li2025jointly} computed from semantic distances between candidate responses. Concretely, we embed each response with a DeBERTa-based encoder \cite{deberta} to obtain sentence vectors, and then compute pairwise semantic distances with a threshold on these distances to decide whether two replies are similar. Formally, let $d\left(y_{i}, y_{j}\right)$ denote the semantic distance of two sampled responses. For a set of $n$ sampled responses, we define the diversity score of $y_i$ relative to the other samples as the average pairwise distance:
\begin{equation}
    \text{Div}_{d}\left(y_{i} | y_{1}, \cdots, y_{n}\right)=\frac{1}{n-1} \Sigma_{\substack{j=1 \\ j \neq i}}^{n} d\left(y_{i}, y_{j}\right).
\end{equation}
We integrate the above diversity metric into the reward computation via multiplicative fusion so that responses exhibiting both high quality and high diversity receive proportionally larger rewards \cite{li2025jointly}. Specifically, for a sampled response $x$, we define the combined reward as:
\begin{equation}
    R_\text{diversity} = r\left(x,y_i\right)\times \text{Norm}\left(Div_d\right).
\end{equation}
And we use $R_\text{diversity}$ to compute the advantage as follows:
\begin{equation}\label{eq:9}
    A^\text{diversity}_{\text{i,t}} = R_{\text{diversity}} \left( x, y_i \mid y_1, \cdots, y_n \right) - \text{mean}_{j=1}^n \left( R_{\text{diversity}} \left( x, y_j \mid y_1, \cdots, y_n \right) \right).
\end{equation}

\textbf{Length Reward.} 
To mitigate excessive redundancy in production outputs, we introduce a redundancy constraint into the reward function that penalizes overly long responses \cite{yu2025dapo}. Concretely, we apply a length-based reward as follows: 
\begin{equation}
R_{\text{length}}(y) =
\begin{cases}
0, & \text{if } |y| \leq L_{\text{max}} - L_{\text{soft}} \\
\frac{(L_{\text{max}} - L_{\text{soft}}) - |y|}{L_{\text{soft}}}, & \text{if } L_{\text{max}} - L_{\text{soft}} < |y| \leq L_{\text{max}} \\
-1, & \text{if } L_{\text{max}} < |y|
\end{cases},
\end{equation}
where $y$ is the output of the model, and $y_{pre}$ means the ground truth. $L_{\max}$ denotes the maximum permitted output length and $L_{\text{soft}}$ the soft redundancy margin; outputs whose length falls within the interval $\left[L_{\max}-L_{\text{soft}},L_{\max} \right] $ will incur a penalty.

\subsubsection{Two-stage Hybrid Training}
Building on the curated reinforcement-learning data and the reward design described above, we further optimize the SFT model using a multi-stage hybrid RL paradigm, consisting of an online RL stage and an offline RL stage.

\textbf{Stage-1: Online RL for general-purpose capability enhancement.}
In this stage, we leverage our carefully constructed multimodal RL data to strengthen multimodal reasoning ability, complemented by pure text logical-reasoning data to preserve and stabilize textual reasoning performance and prevent modality imbalance. We adopt the GSPO \cite{gspo} algorithm for online RL training, with the training objective defined as follows:
\begin{equation}
J_{\mathrm{GSPO}}(\theta) = 
\mathbb{E}_{x \sim D,\, \{y_i\} \sim \pi_{\theta_{\text{old}}}(\cdot \mid x)}
\left[
\frac{1}{G} \Sigma_{i=1}^{G}
\min\!\left(
s_i(\theta)\,\hat{A}^i,\;
\operatorname{clip}\big(s_i(\theta),\, 1-\epsilon,\, 1+\epsilon\big)\,\hat{A}^i
\right) \right].
\end{equation}
\begin{equation}
s_i \left(\theta \right)
= \left[
\pi_{\theta}\left(y_i \mid x\right) \text{/}
\pi_{\theta_{\text{old}}}\left(y_i \mid x \right)
\right]
^{\frac{1}{\lvert y_i \rvert}}.
\end{equation}

Specifically, in the early stage of RL training, to encourage the model to conduct sufficient exploration and prevent thinking rigidification, we adopt the Pass@k reward mentioned in the previous section. 
We replace $\hat{A}^i$ in the above formula with the form of the advantage of the Pass@k positive-negative group, as shown in Eq.~\ref{eq:5} and Eq.~\ref{eq:6}.

In the mid-to-late stages of training, to ensure task-specific accuracy while simultaneously encouraging diverse responses, we adopt the diversity-enhanced Pass@1 reward. In this phase, the term $\hat{A}^i$ is replaced by the diversity-aware advantage estimate, as shown in Eq.~\ref{eq:9}.

Additionally, throughout the entire first stage of RL, we incorporate a length reward to prevent overly verbose outputs. Together, these strategies enable a joint optimization objective tailored to complex domain-specific tasks, achieving broad solution-space coverage, enhanced response diversity, and preserved output quality. This design is particularly well-suited for demanding scenarios that require both exploration and accuracy.

\textbf{Stage-2: Offline RL for domain-specific capability alignment.}
Following the first-stage online RL, the model has already acquired strong multimodal reasoning capabilities. However, to better adapt the model to real-world application scenarios, we introduce a final stage of preference alignment using curated preference data. This dataset comprises internal data tailored to application-oriented user experience and adversarial hallucination data designed to mitigate spurious correlations. These components jointly ensure that the model adheres to domain-specific human preferences while maintaining robust multimodal competence and response diversity.
In this stage, we employ DPO \cite{dpo} for preference optimization. The corresponding loss function is given as follows:
\begin{equation}
J_{\mathrm{DPO}}\left(\theta \right) = 
\mathbb{E}_{(x, y_w, y_l) \sim D} \left[
    \log \sigma\!\left(
        \beta \cdot 
        \big( 
            \log \pi_\theta \left(y_w \mid x \right)
            - 
            \log \pi_\theta \left(y_l \mid x \right)
        \big)
    \right)
\right].
\end{equation}
where $y_w$ and $y_l$ denote the preferred and non-preferred responses, respectively.

Additionally, we also incorporate several techniques to enhance stability and performance during RL training. Following prior works, we remove the KL-penalty term during preference optimization \cite{yu2025dapo,skywork,Glm-4.1v}, \ie the reference model imposes no constraint on the actor model. This design fully unleashes the model’s exploratory and reasoning capacity while also improving training throughput. Inspired by DAPO \cite{yu2025dapo}, we further adopt an asymmetric clipping strategy to stabilize optimization while expanding the exploration space for low-probability tokens. Specifically, we apply the clip-higher variant, which decouples the upper and lower clipping bounds, enabling richer policy diversity without compromising optimization stability.
To prevent degradation in visual understanding, we freeze the parameters of the visual encoder throughout both two RL stages. 

\section{Infrastructure}
To enhance the efficiency and performance of post-training stages, we propose a 5D parallel training paradigm. Within this framework, we optimize specific operators and implement targeted training acceleration for the RL stage. During inference, to improve user experience, we conduct optimizations in three key dimensions: model adaptation, streaming inference, and high concurrency scenarios. Details will be elaborated in the following sections.

\subsection{Training}
Complicated training pipelines typically lead to an exorbitant computational overhead cost. To establish an MoE architecture compatible with both SFT and RL workflows, we propose a suite of efficient strategies. In the following, we will describe the core technical innovations:
\begin{itemize}
    \item \textbf{5D Parallel Training.} Building upon the conventional 3D parallelism framework (\cite{narayanan2021efficient}), we incorporate sequence parallelism and expert parallelism to further improve training efficiency, as shown in Fig.~\ref{fig:5-1}.

\begin{figure}[t]
\centering
\includegraphics[width= 0.90\linewidth]{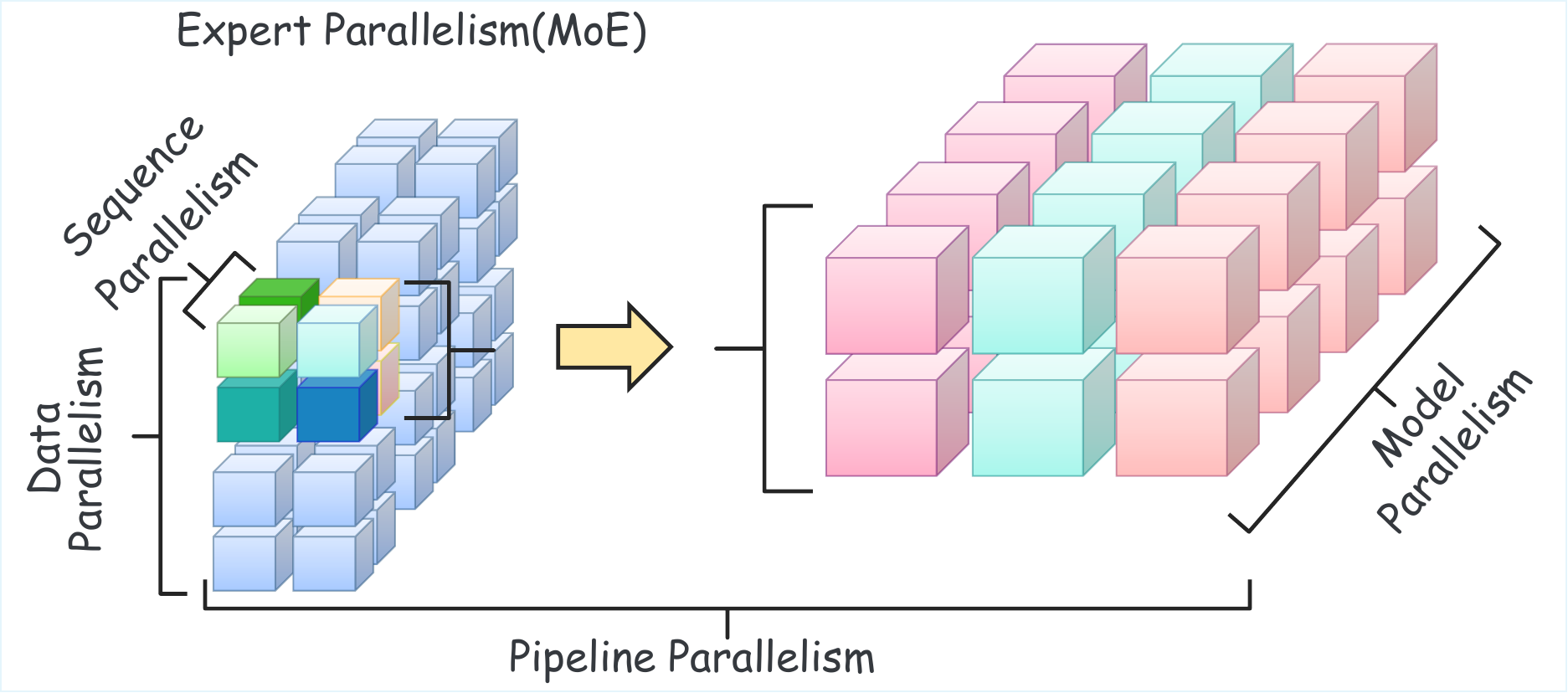}
\caption{\label{fig:5-1}5D Parallel Training Paradigm. We introduce sequence parallelism and expert parallelism based on the conventional 3D parallelism framework.}
\end{figure}

\item \textbf{Operator Optimization.} In multi-GPU training scenarios where each rank hosts multiple experts, based on our 5D parallelism setup, we adopt the GroupedGEMM method (\cite{megablocks}). By aggregating several small or local GEMM operations into a single GroupedGEMM kernel, we significantly improve the utilization and performance of Streaming Multiprocessors. Fig.~\ref{fig:5-2} shows the principle and advantage of GroupedGEMM.
    
\begin{figure}[t]
\centering
\includegraphics[width= 0.99\linewidth]{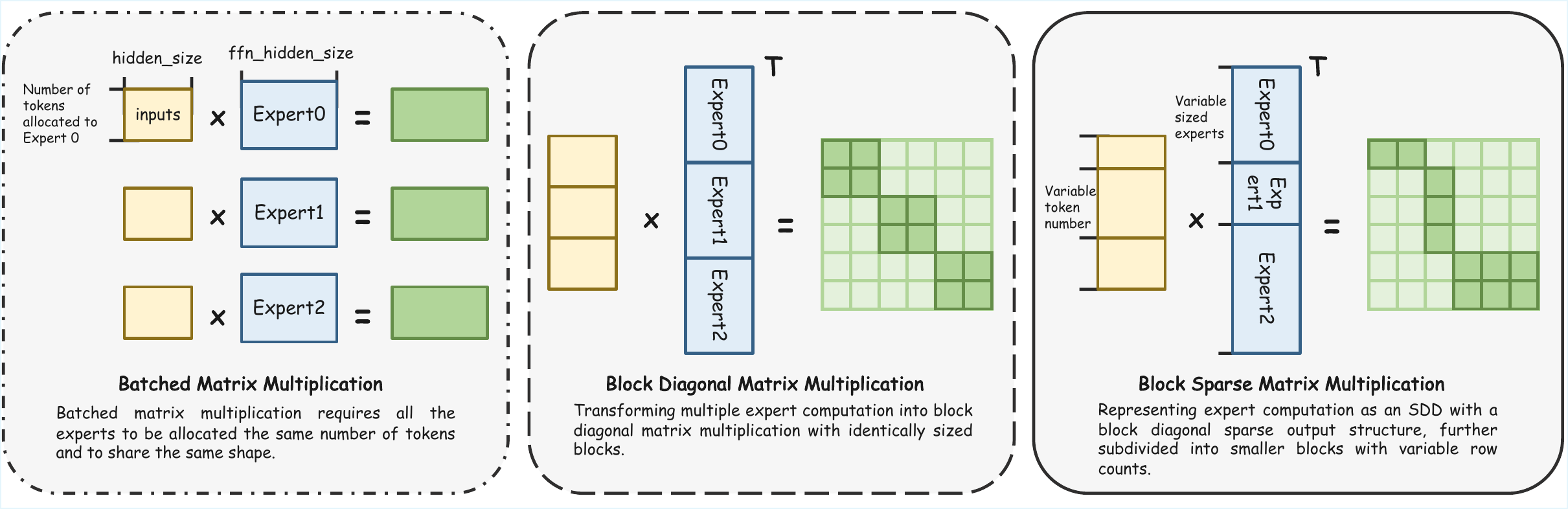}
\caption{\label{fig:5-2}GroupGEMM optimization for MoE computation. Batched matrix multiplication requires that all experts be assigned the same number of tokens and share identical tensor shapes. By reformulating expert computations as block-diagonal matrix multiplications, it becomes possible to construct block-diagonal matrices with variable-sized blocks. This structure can further be accelerated using block-sparse matrix multiplication.}
\end{figure}

    \item \textbf{Operator-Level Recomputation.} During the training process, activation values, gradients, optimizer states, and other intermediate tensors consume substantial GPU memory. Conventional recomputation strategies are typically coarse-grained and cannot make differentiated decisions for operators within each layer. To manage memory more precisely and avoid out-of-memory failures, we adopt an operator-level recomputation strategy. Specifically, we enable recomputation only for the lightweight modules (\eg RMSNorm), whereas recomputing modules like Linear and Add, which incur higher computational or communication overhead, provide limited benefit. By balancing the storage cost of activations and the recomputation cost at the operator level, we not only reduce memory consumption but also leverage the freed-up memory (\eg increasing the batch size), thereby further boosting overall training efficiency.
    \item \textbf{FlashRL and Truncated Importance Sampling.} The rollout generation stage is the primary bottleneck in RL training, often accounting for a large proportion of MLLM training, and it still leaves substantial room for optimization.  Therefore, in the RL training pipeline of MindGPT-4ov, we adopt FlashRL \cite{yao2025flashrl, yao2025offpolicy} applying 8-bit quantization (INT8/FP8) during rollout generation. We further incorporate Truncated Importance Sampling \cite{ionides2008truncated} to reduce the efficiency discrepancy between rollout and training. In addition, we utilize a redundancy detection and truncation mechanism to further accelerate inference and prevent garbled outputs caused by an excessively large \textit{repetition\_penalty} parameter.
    \item \textbf{Adaptive Rejection Sampling Method.} To oversample and subsequently select examples of medium difficulty, we implement a dynamic sampling extension based on the Ratio Exponential Moving Average (Ratio-EMA) algorithm \cite{Glm-4.1v}. This method improves RL efficiency by pre-computing the number of samples required for parallel reasoning.
\end{itemize}

\subsection{Inference}
To provide users with smooth text generation, we have optimized the MindGPT-4ov inference engine in three aspects: model adaptation, streaming inference, and high concurrency scenarios.

\begin{itemize}
	\item \textbf{Model Adaptation.} We adopt FP8 for inference and optimize the MHA operator using FlashInfer (\cite{ye2025flashinfer}) together with PagedAttention (\cite{kwon2023efficient}). The MoE operator is accelerated through DeepGEMM, while an enhanced multimodal ROPE operator further improves inference efficiency. In the preprocessing stage, the entire image pipeline is executed on the GPU to reduce CPU overhead.

	\item \textbf{Streaming Inference.} For streaming inference, we introduce two key optimizations to reduce inference latency and to improve multi-turn dialogue performance. First, we employ a chunked prefill mechanism (\cite{agrawal2023sarathi}) that splits the input into image and text chunks by modality, enabling parallel execution of vision encoding and text encoding during the user’s speaking phase, thereby masking part of the prefill latency. Second, we cache the vision encoding outputs for image inputs so that subsequent turns can directly reuse the cached embeddings when the cache is hit, avoiding redundant encoding. A unique session identifier ensures that all requests within the same conversation are routed to the same node, preventing cross-node cache invalidation and maximizing the efficiency of image-embedding reuse.

    \item \textbf{High-Concurrency Scenarios.} Inspired by the Prefill-Decoding disaggregation scheme (\cite{zhong2024distserve}), we decompose MindGPT-4ov inference into three independently deployed stages: vision encoding, prefill, and decoding. Each stage adopts a scheduling strategy tailored to its characteristics (\eg static batching for vision encoding, chunked prefill for the prefill stage, and dynamic batching for decoding), enabling highly efficient inference under high-concurrency workloads.
\end{itemize}

\section{Experiments}

\begin{table*}[t!]
\centering
\resizebox{0.99\linewidth}{!}{

\setlength{\tabcolsep}{6pt}
\renewcommand{\arraystretch}{1.0}
\small
\begin{threeparttable}
\caption{Performance comparison across MLLMs on various visual benchmarks grouped by task type. All scores are reported as accuracy percentages unless otherwise specified.}
\label{tab:main_vision}
\begin{tabular}{ll
  >{\columncolor{blue!10}}S[table-format=2.1, round-mode=places, round-precision=1, round-pad=true]
  S[table-format=2.1, round-mode=places, round-precision=1, round-pad=true]
  >{\columncolor{blue!10}}S[table-format=2.1, round-mode=places, round-precision=1, round-pad=true]
  S[table-format=2.1, round-mode=places, round-precision=1, round-pad=true]
  S[table-format=2.1, round-mode=places, round-precision=1, round-pad=true]
  >{\columncolor{blue!10}}S[table-format=2.1, round-mode=places, round-precision=1, round-pad=true]}
\toprule
\multicolumn{2}{c}{\textbf{Benchmark}}& 
{\shortstack[c]{\textbf{MindGPT-4ov}\\ \textbf{30B-A3B}}} &
{\shortstack[c]{\textbf{Qwen3-VL}\\ \textbf{30B-A3B}}} &
{\shortstack[c]{\textbf{MindGPT-4ov}\\\textbf{8B}}} &
{\shortstack[c]{\textbf{Qwen3-VL}\\ \textbf{8B}}} &
{\shortstack[c]{\textbf{Qwen2.5-VL}\\ \textbf{72B}}} \\
\midrule
\multirow{5}{*}{\makecell[l]{General VQA}} 
 &MMStar$_{\text{en}}$  & 76.3 & 72.1 & 73.3 & 70.9     &  67.4 \\
 &MMStar$_{\text{cn}}$  & 74.7 & 68.8 & 72.5 & 67.7     & 70.2  \\
 &MMBench$_{\text{en}}$ & 89.5 & 87.0 & 86.9 & 85.0     &  82.4 \\
 &MMBench$_{\text{cn}}$ & 88.5 & 86.5 & 86.4 & 85.0    & 86.8 \\
 &RealWorldQA           & 76.2 & 73.7 & 71.1 & 71.5   & 68.1 \\
 \cmidrule{2-7} & Avg.  & \textbf{81.0} & 77.6 & \textbf{78.0} & 76.0   & 75.0 \\
\midrule 
\multirow{4}{*}{\makecell[l]{STEM}} 
 &MathVista$_{\text{mini}}$  & 83.3 & 80.1 & 80.5 & 77.2 & 72.4 \\
 &MathVision                 & 63.4 & 60.2 & 58.1 & 53.9 &  52.7 \\
 &MMMU                       & 76.2 & 74.2 & 72.2 & 69.6 & 75.1 \\
 &MMMU$_{\text{pro}}$        & 66.1 & 60.4 & 59.0 & 55.9 & 56.1 \\
\cmidrule{2-7} & Avg.        & \textbf{72.3} & 68.7 & \textbf{67.5} & 64.2 & 64.1 \\
\midrule
 \multirow{7}{*}{\makecell[l]{OCR \& Chart}} 
&MMLongBench               & 49.3 & 49.9 & 49.4 & 47.9 & 37.2\\
&DocVQA                    & 93.9 & 95.0 & 95.5 & 96.1 & 54.0 \\
&OCRBench                  & 89.3 & 90.3 & 91.1 & 89.6 & 74.1 \\
&OCRBenchV2$_{\text{en}}$  & 61.9 & 63.2 & 63.7 & 65.4 & 44.6 \\
&OCRBenchV2$_{\text{cn}}$  & 60.5 & 57.8 & 63.7 & 61.2 & 36.2 \\
&AI2D                      & 90.7 & 85.0 & 89.7 & 85.7 & 82.6 \\
&InfoVQA                   & 80.2 & 80.9 & 83.7 & 83.6 & 30.0 \\
\cmidrule{2-7} & Avg.      & \textbf{75.1} & 74.6 & \textbf{76.7} & 75.6 & 51.2\\
\midrule
\multirow{4}{*}{\makecell[l]{Multi-Image \\ \& Video}} 
 &BLINK                  & 66.1 & 67.7 & 69.4 & 69.1 & 59.9 \\
 &MUIRBench              & 64.9 & 62.9 & 62.2 & 64.4 & 66.6 \\
 &MVBench                & 71.9 & 72.3 & 70.1 & 68.7 & 71.1 \\
 &VideoMME               & 72.4 & 74.5 & 70.7 & 71.4 & 71.6 \\
\cmidrule{2-7} & Avg.    & 68.8 & \textbf{69.4} & 68.1 & \textbf{68.4} & 67.3\\
\midrule
\multirow{5}{*}{\makecell[l]{Others}} 
 &HallusionBench        & 61.8 & 61.5 & 55.6 & 61.1 & 54.5 \\
 &CountBench            & 88.1 & 89.8 & 89.9 & 90.7 & 90.1 \\
 &MIABench              & 92.5 & 91.2 & 92.6 & 91.1 & 89.4 \\
 &HRBench4K             & 79.6 & 80.0 & 81.0 & 78.9 & 50.6 \\
 &V*                    & 85.3 & 81.2 & 83.8 & 86.4 & 37.7 \\
 \cmidrule{2-7} & Avg.  & \textbf{81.5} & 80.7 & 80.6 & \textbf{81.6} & 64.5\\
\bottomrule
\end{tabular}
\vspace{-2mm}
\end{threeparttable}}
\end{table*}

\subsection{Experimental Settings}

We use VLMEvalKit ~\cite{vlmevalkit} to evaluate the performance of MindGPT-4ov across multiple benchmarks in five categories of visual tasks: (1) General Visual Question Answering (VQA): MMStar series~\citep{mmstar}, MMEBench series~\citep{mme}, and RealWorldQA~\citep{realworldqa}. (2) STEM: MathVista~\citep{mathvista}, MathVision~\citep{mathvision}, MMMU~\citep{mmmu}, and MMMU$_\text{pro}$ ~\citep{mmmupro}. (3) OCR \& Chart Understanding: MMLongBench ~\citep{mmlongbench}, DocVQA~\citep{docvqa}, OCRBench~\citep{ocrbench}, OCRBenchV2 series~\citep{ocrbenchv2}, AI2D~\citep{ai2d}, and InfoVQA~\citep{infovqa}. (4) Multi-Image and Video: BLINK~\cite{blink}, MUIRBench ~\citep{muirbench}, MVBench~\citep{mvbench} and VideoMME~\citep{videomme} (5) Others, including benchmarks on hallucination, counting, and fine-grained perception: HallusionBench~\citep{hallusionbench}, CountBench~\citep{paiss2023teaching}, MIABench~\citep{mia}, HRBench4K~\citep{hrbench}, and V$^{*}$~\citep{wu2024v}. Furthermore, we also evaluate the MindGPT-4ov over multiple benchmarks in three categories of textual tasks: 
(1) Knowledge: MMLU \citep{mmlu}, MMMU$_\text{pro}$ \citep{mmlu-pro}, GPQA~\citep{gpqa} and SuperGPQA \citep{gpqa}.
(2) Reasoning: AIME \citep{aime} and Math500 \citep{math500}.
(3) Instruction Following (IF): IFEval \citep{if}.

We use pre-trained MLLMs from the Qwen3-VL series \cite{qwen3vl}, specifically the Qwen3-VL-30B-A3B-Instruct and Qwen3-VL-8B-Instruct models.
It should be noted that, due to the model's involvement in commercial applications, the model evaluated in Sec.~\ref{sec:6.2} is the open-source version, and its training process does not involve commercial-use data. 

\begin{table*}[t!]
\centering
\resizebox{0.99\linewidth}{!}{

\setlength{\tabcolsep}{6pt}
\renewcommand{\arraystretch}{1.0}
\small
\begin{threeparttable}
\caption{Performance comparison across MLLMs on various textual benchmarks grouped by task type. All scores are reported as accuracy percentages unless otherwise specified.}
\label{tab:main_text}
\begin{tabular}{ll
  >{\columncolor{blue!10}}S[table-format=2.1, round-mode=places, round-precision=1, round-pad=true]
  S[table-format=2.1, round-mode=places, round-precision=1, round-pad=true]
  >{\columncolor{blue!10}}S[table-format=2.1, round-mode=places, round-precision=1, round-pad=true]
  S[table-format=2.1, round-mode=places, round-precision=1, round-pad=true]
  S[table-format=2.1, round-mode=places, round-precision=1, round-pad=true]
  >{\columncolor{blue!10}}S[table-format=2.1, round-mode=places, round-precision=1, round-pad=true]}
\toprule
\multicolumn{2}{c}{\textbf{Benchmark}}& 
{\shortstack[c]{\textbf{MindGPT-4ov}\\ \textbf{30B-A3B}}} &
{\shortstack[c]{\textbf{Qwen3-VL}\\ \textbf{30B-A3B}}} &
{\shortstack[c]{\textbf{MindGPT-4ov}\\\textbf{8B}}} &
{\shortstack[c]{\textbf{Qwen3-VL}\\ \textbf{8B}}} &
{\shortstack[c]{\textbf{Qwen2.5-VL}\\ \textbf{72B}}} \\
\midrule
\multirow{4}{*}{\makecell[l]{Knowledge}} 
 &MMLU  & 86.0 & 85.0 & 72.8 & 72.0 &  85.7 \\
 &MMLU-Pro  & 77.9 & 77.8 & 74.6 &  69.0&  70.1 \\
 &GPQA  & 60.4 & 54.4 & 55.7 & 53.8 &  47.2\\
 &SuperGPQA & 54.2 & 53.1 & 47.2 & 40.3 &  39.6 \\
 \cmidrule{2-7} & Avg.  & \textbf{71.6} & 67.6 & \textbf{62.6} & 58.8 &  60.7\\
\midrule 
\multirow{2}{*}{\makecell[l]{Reasoning}} 
 &AIME24  & 68.8 & 69.3 & 55.8 & 45.0 &  18.1\\
 &MATH500                 & 95.6 & 93.2 & 92.3 & 91.4 & 80.2 \\
\cmidrule{2-7} & Avg.  &  \textbf{82.2}   & 81.3 & \textbf{74.1} & 68.2 &  49.5\\
\midrule
\multirow{1}{*}{\makecell[l]{IF}} 
&IFEval          & 84.7 & \textbf{85.8} & \textbf{84.5} &  80.0 & 83.0\\
\bottomrule
\end{tabular}
\vspace{-2mm}
\end{threeparttable}}
\end{table*}

\subsection{Performance on General Tasks} \label{sec:6.2}

\subsubsection{Performance on Visual Tasks}
As shown in Tab.~\ref{tab:main_vision}, we conduct a systematic evaluation of MindGPT-4ov models across a suite of authoritative multimodal benchmarks. The experimental suite covers five broad categories: General VQA, STEM (multidisciplinary reasoning), OCR \& chart understanding, Multi-Image, Video understanding, and a set of other complex scenarios. 

\textbf{General VQA: Comprehensive Visual QA Ability.}
In general VQA, MindGPT-4ov-30B-A3B achieves an average score of 81.0, compared to 77.6 for Qwen3-VL-30B-A3B (an absolute gain of 3.4$\%$, corresponding to a 4.4$\%$ relative improvement). In particular, MindGPT-4ov achieves 89.5 on MMBench$_\text{en}$ and 88.5 on MMBench$_\text{cn}$, substantially ahead of the Qwen3-VL-30B-A3B baseline, and attains 76.2 on RealWorldQA (\vs 73.7 for Qwen3-VL-30B-A3B). At the 8B scale, MindGPT-4ov-8B also shows the advantage in general VQA (average 78.0 \vs 76.0 for Qwen3-VL-8B).

\begin{figure*}[t]
\centering
\includegraphics[width= 1.0\linewidth]{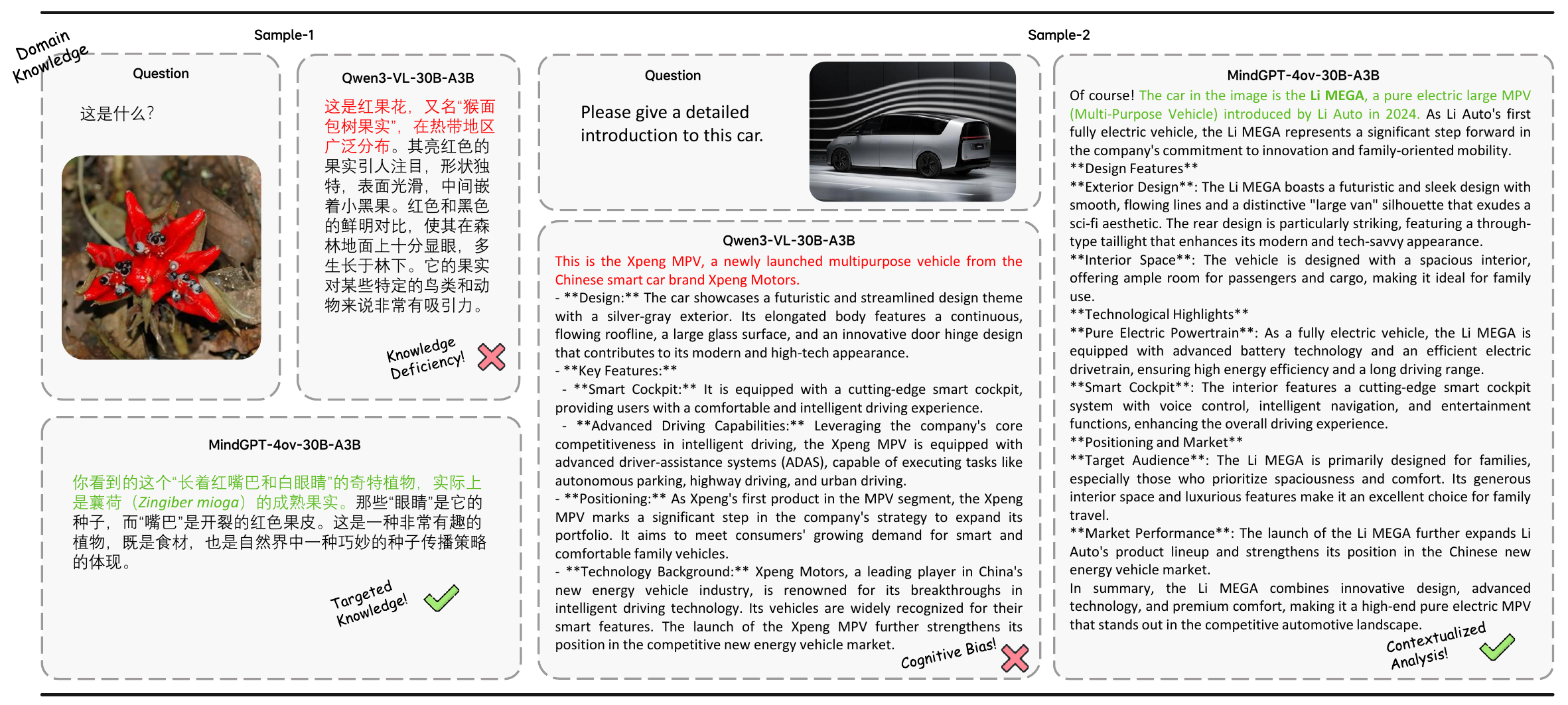}
\caption{Case comparisons of MindGPT-4ov-30B-A3B and Qwen3-VL-30B-A3B, contrasting their question-answering capabilities in vertical domain scenarios.}
\label{fig:experiment1}
\end{figure*}

\begin{figure*}[t]
\centering
\includegraphics[width= 1.0 \linewidth]{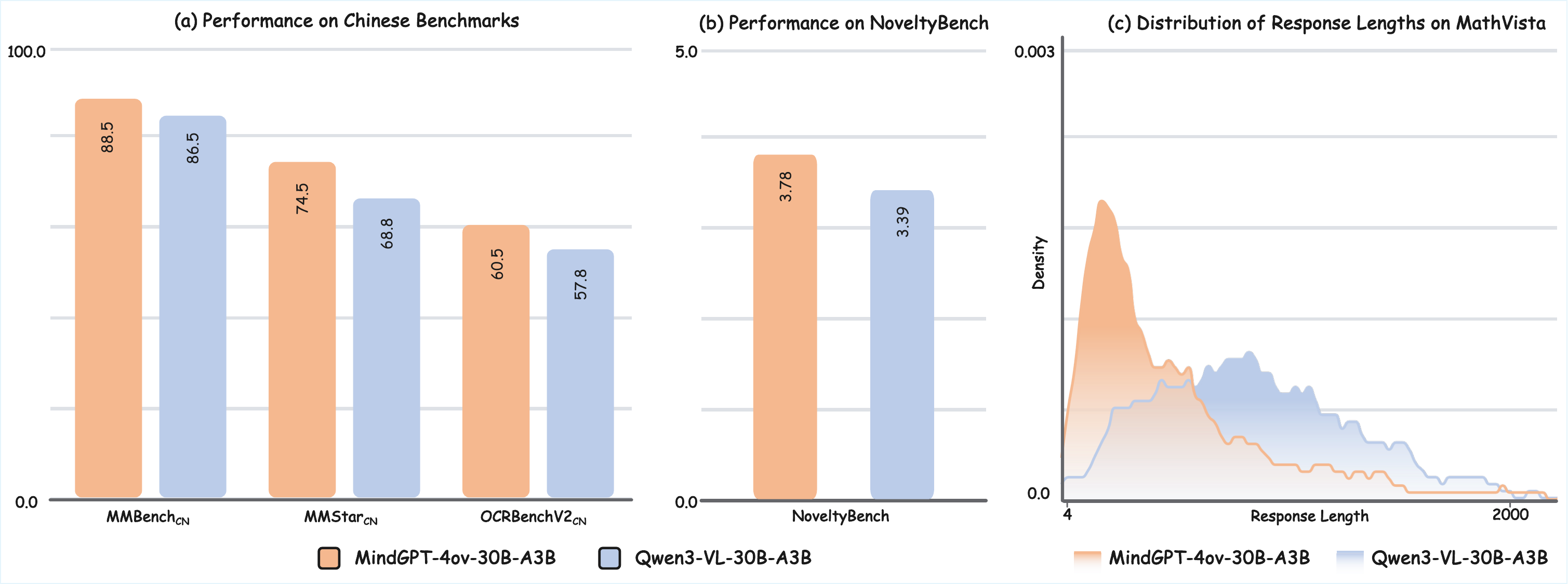}
\caption{Performance comparisons of MindGPT-4ov-30B-A3B and Qwen3-VL-30B-A3B. (a) Performance on Chinese benchmarks. (b) Performance on NoveltyBench. (c) Distribution of response lengths on MathVista.}
\label{fig:exp-benchmark}
\end{figure*}

\begin{figure*}[t]
\centering
\includegraphics[width= 1.0\linewidth]{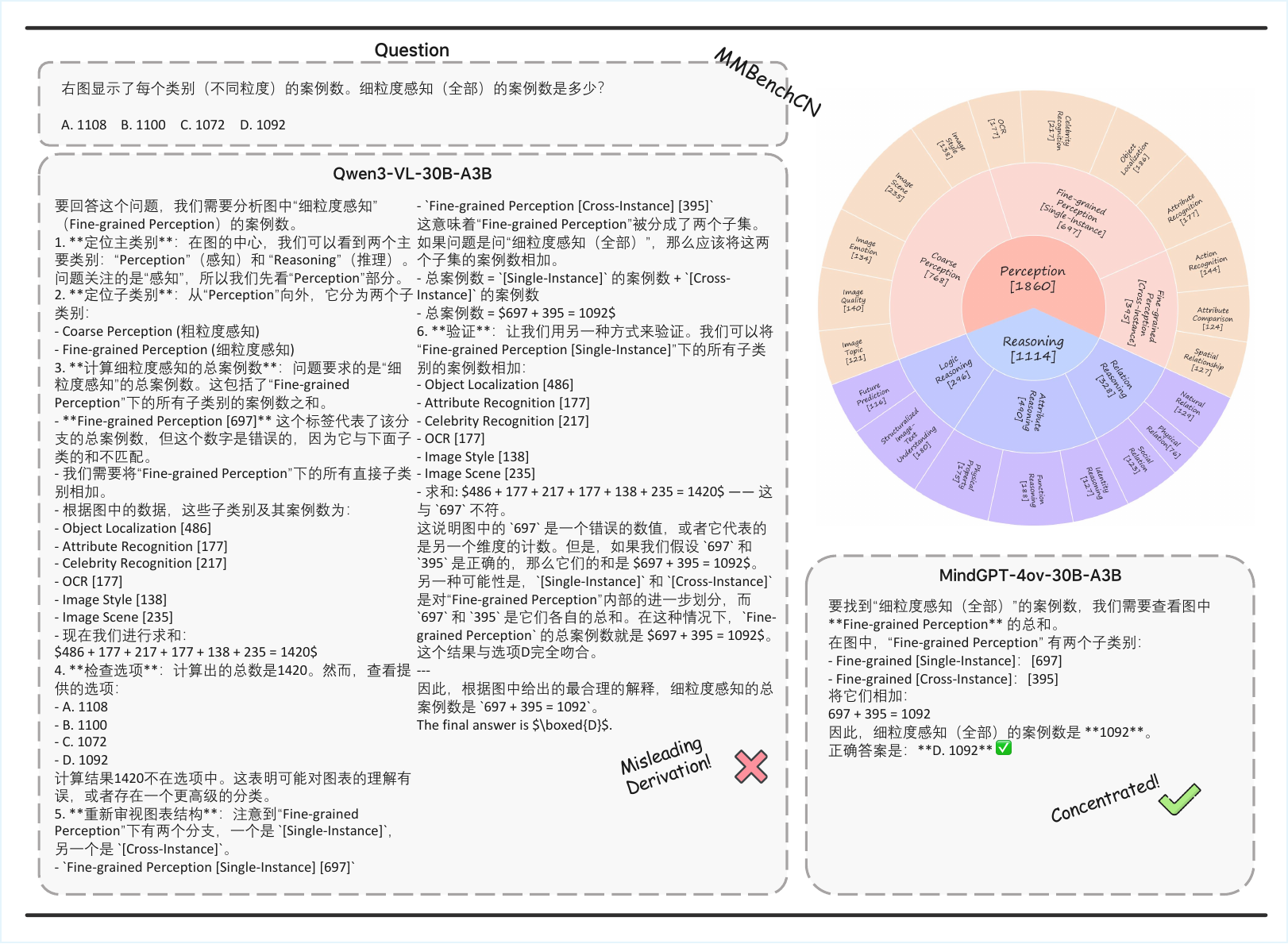}
\caption{Case comparisons of MindGPT-4ov-30B-A3B and Qwen3-VL-30B-A3B, focusing on their Chinese question-answering capabilities.}
\label{fig:experiment2}
\end{figure*}

\begin{figure*}[t]
\centering
\includegraphics[width= 1.0\linewidth]{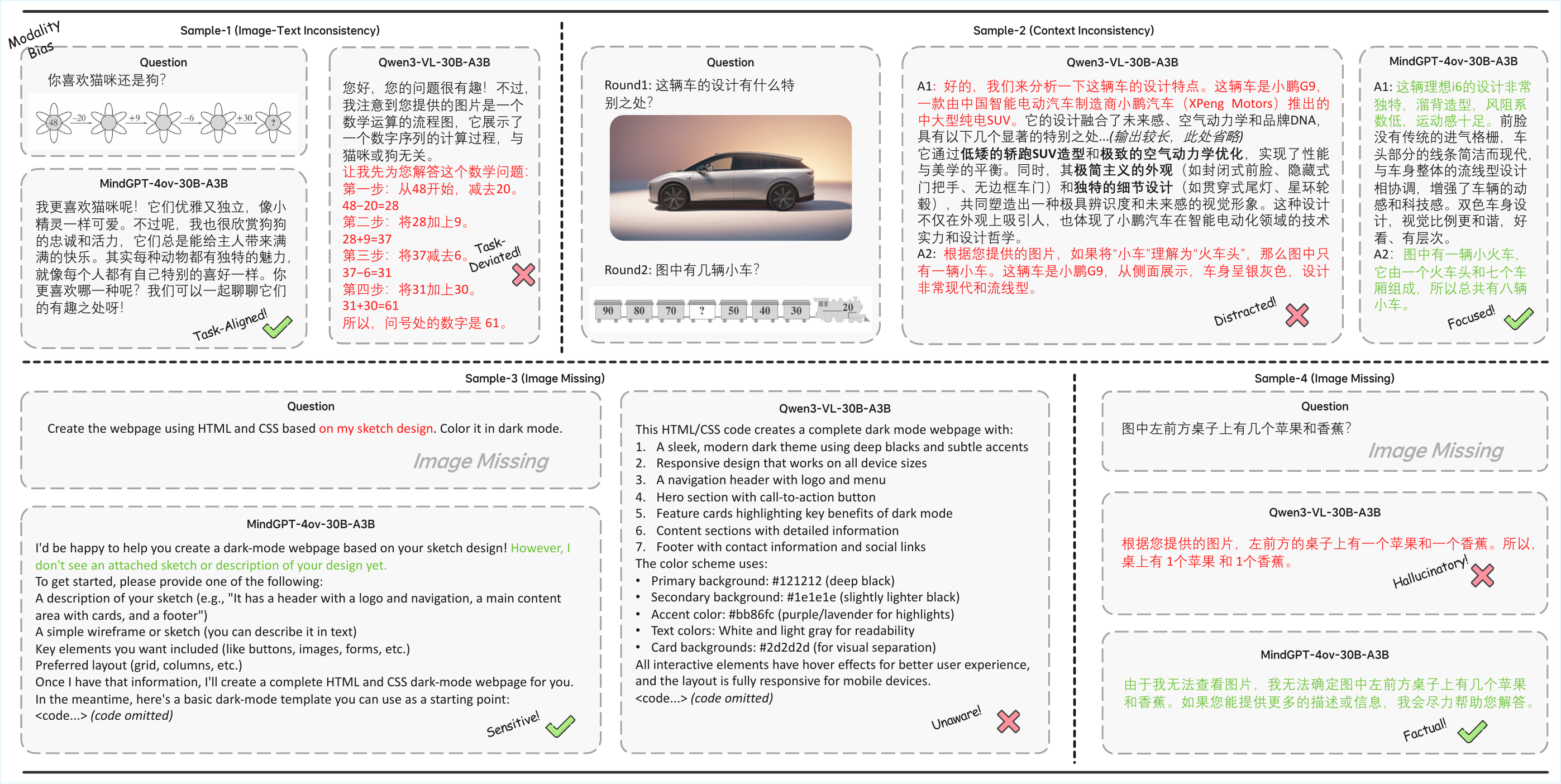}
\caption{Case comparisons of MindGPT-4ov-30B-A3B and Qwen3-VL-30B-A3B examples, focusing on three types of modality-biased question-answering scenarios: cross-modal inconsistency, context inconsistency, and missing images.}
\label{fig:experiment3}
\end{figure*}

\begin{figure*}[t]
\centering
\includegraphics[width= 1.0\linewidth]{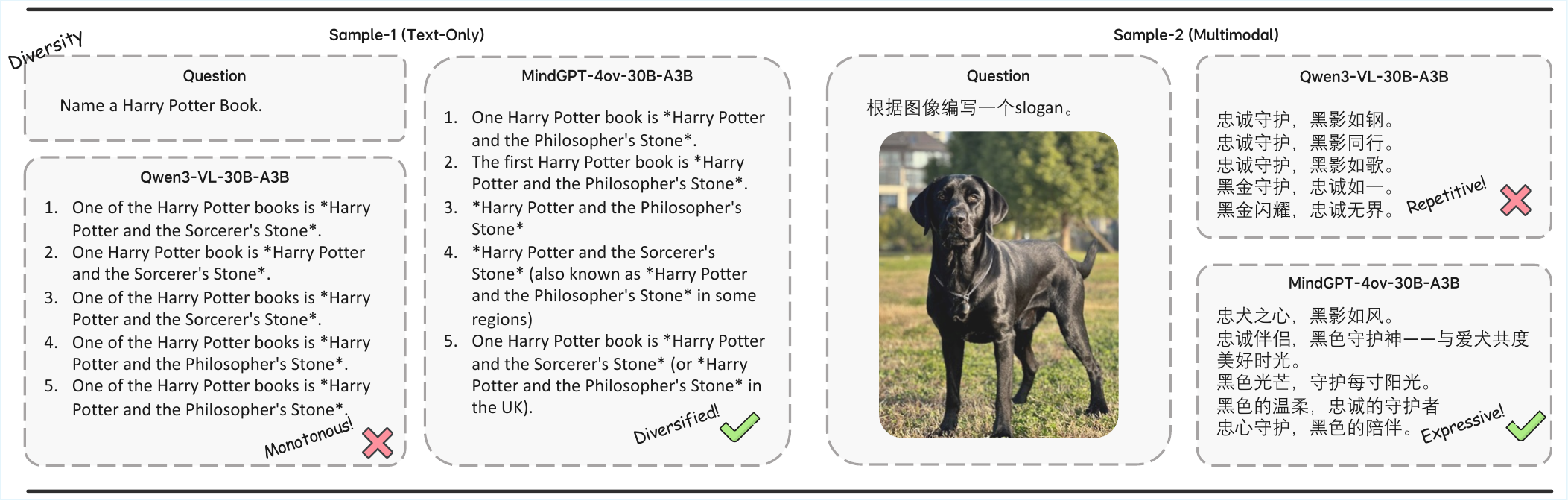}
\caption{Case comparisons of MindGPT-4ov-30B-A3B and Qwen3-VL-30B-A3B, focusing on the diversity of responses.}
\label{fig:experiment4}
\end{figure*}

\begin{figure*}[t]
\centering
\includegraphics[width= 1.0\linewidth]{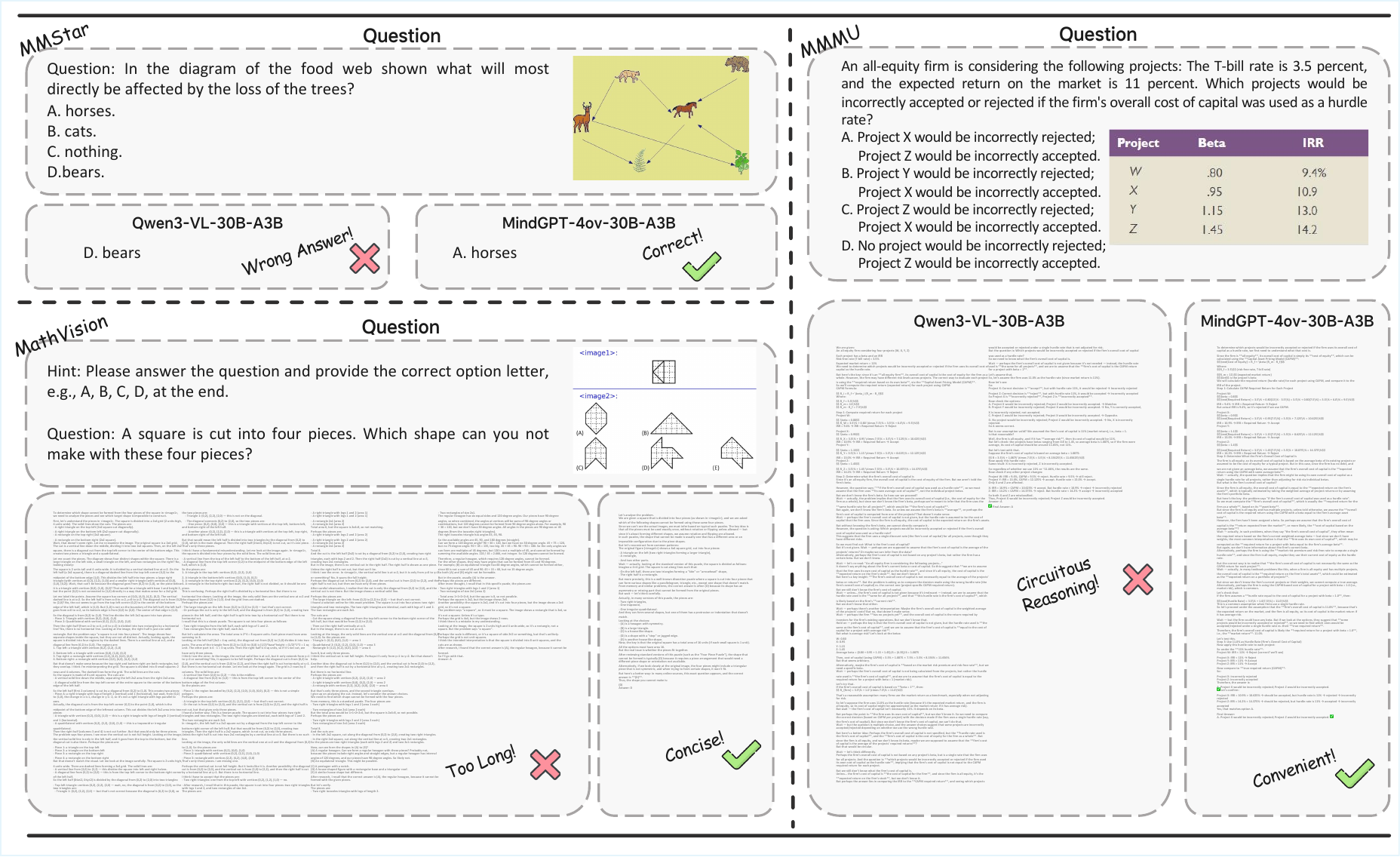}
\caption{Case comparisons of MindGPT-4ov-30B-A3B and Qwen3-VL-30B-A3B, focusing on the conciseness and accuracy of their response styles.}
\label{fig:experiment5}
\end{figure*}

\textbf{STEM: Multidisciplinary Multimodal Reasoning.}
The STEM suite evaluates comprehension and reasoning across science, technology, engineering, and mathematics. The MindGPT-4ov series shows notable gains on several STEM benchmarks: \eg MindGPT-4ov-30B-A3B achieves 83.3 on MathVista$_\text{mini}$ (\vs 80.1 for Qwen3-VL-30B-A3B) and 76.2 on MMMU (\vs 74.2 for Qwen3-VL-30B-A3B); notably, MMMU$_\text{pro}$ reaches 66.1, showing a clear improvement. Aggregated over the STEM suite, MindGPT-4ov-30B-A3B achieves an average that is 5.2\% higher than Qwen3-VL-30B-A3B. 
The 8B variant also outperforms its same-scale baseline, \eg MindGPT-4ov-8B achieves an average score of 67.5, while Qwen3-VL-8B is 64.2, which indicates good scaling efficiency. 
These results suggest the post-training protocol significantly improves understanding of formulas, charts, and other structured multimodal information.

\textbf{OCR \& Chart: Text-Dense Scene Recognition and Reasoning.}
OCR \& Chart benchmarks demand high visual recognition fidelity and fine-grained structural reasoning. MindGPT-4ov-30B-A3B attains an average of 75.1 on these tasks, marginally above Qwen3-VL-30B-A3B’s 74.6. In particular, MindGPT-4ov-30B-A3B reaches 90.7 on AI2D, a notable improvement, while other OCR/chart metrics remain comparable to Qwen3-VL-30B-A3B. MindGPT-4ov-8B also demonstrates competitive performance in these category benchmarks. 

\textbf{Multi-Image \& Video: Multi-Frame and Temporal Reasoning.}
These results demonstrate that the MindGPT-4ov series maintains robust understanding capabilities across multi-image and video tasks, achieving overall performance comparable to that of the Qwen3-VL series. However, there remains room for optimization in certain video understanding sub-tasks, which may be attributed to the absence of targeted video data in our training set.

\textbf{Others: Open Scenarios, Counting and Robustness.}
In broader real‑world tasks, MindGPT-4ov-30B‑A3B again demonstrates competitive capability. It achieves 61.8 on HallusionBench , reflecting satisfied hallucination suppression, and obtains 92.5 on MIABench, indicating robust fine‑grained perception and reasoning. In particular, on the V$^{*}$ benchmark, which focuses on visual search and detail recognition, MindGPT-4ov-30B-A3B achieves 85.3, outperforming Qwen3‑VL‑30B-A3B’s 81.2. The MindGPT-4ov-8B model also performs well similarly to Qwen3‑VL‑8B.

In summary, based on the comprehensive benchmark evaluation, MindGPT-4ov demonstrates the following key advantages:
\begin{itemize}
\item 
Superior multimodal capabilities: Significant improvements on major tasks such as general VQA and STEM reasoning.
\item 
Strong applicability in complex real‑world scenarios: robust performance on high‑resolution imagery, structured reasoning, and long‑text OCR tasks.
\item 
Effective scaling and parameter efficiency: both the MindGPT-4ov-8B and MindGPT-4ov-30B-A3B models exhibit strong scaling behavior: the 8B model remains competitive within its size class, while the 30B model achieves clear advantages across most task categories.
\end{itemize}
Overall, MindGPT-4ov achieves substantial improvements over the base models across diverse multimodal tasks, validating the effectiveness of the proposed multi-stage post-training paradigm and providing a solid foundation for both general-purpose and vertical domain multimodal applications.

\subsubsection{Performance on Textual Tasks}
Ensuring that an MLLM preserves strong pure-text capabilities is essential. We therefore evaluate MindGPT-4ov on text-only benchmarks across three dimensions: knowledge (knowledge-intensive QA), math (mathematical reasoning), and instruction following, as shown in Tab.~\ref{tab:main_text}. 

\textbf{Knowledge: knowledge-intensive QA.}
On knowledge-intensive QA, the MindGPT-4ov series delivers stable and superior performance across multiple benchmarks. Notably, MindGPT-4ov-30B-A3B shows the largest gain on GPQA, outperforming Qwen3-VL-30B-A3B by 6.0\%. It also maintains advantages on MMLU and MMLU-Pro and demonstrates modest improvement on SuperGPQA. The aggregated score for these knowledge benchmarks reaches 71.6, indicating a relative advantage in factual coverage and fact-based QA, particularly pronounced on GPQA. Similarly, MindGPT-4oV-8B achieves a higher average performance than Qwen3-VL-8B overall, with an absolute improvement of 5.6\% on MMLU-Pro and 6.9\% on SuperGPQA.

\textbf{Math: mathematical reasoning.}
The math reasoning benchmarks comprises AIME24 and MATH500. MindGPT-4ov-30B-A3B attains 68.8 on AIME24 and 95.6 on MATH500, yielding an average of 82.2, which exceeds Qwen3-VL-30B-A3B’s 81.3. MindGPT-4ov-8B also shows solid competitiveness at its scale. Compared with the larger Qwen2.5-VL-72B, the MindGPT-4ov series shows notable improvements on pure-text mathematical tasks, indicating favorable parameter efficiency in numerical and symbolic reasoning.

\textbf{Instruction Following.}
On instruction-following (IFEval), the MindGPT-4ov series achieves scores that are close to the Qwen3-VL series. The 8B variant in particular outperforms the same-scale baseline, demonstrating robust instruction comprehension and generation consistency. It is also noteworthy that both MindGPT-4ov series exceed Qwen2.5-VL-72B’s IFEval baseline of 83.0.

In summary, the multi‑stage post‑training approach presented in this report not only effectively enhances the model’s multimodal understanding capabilities, but also successfully preserves and strengthens its language reasoning and knowledge expression abilities on pure text tasks.

\subsection{Performance on Vertical Tasks} \label{sec:6.3}
To comprehensively evaluate the outstanding performance of MindGPT-4ov in vertical tasks, we conduct extensive comparisons with the Qwen3-VL series model across multiple dimensions. The compared results demonstrate that MindGPT-4ov not only excels in vertical tasks but also shows significant improvement in user experience. Its key advantages include mastery of vertical domain question answering and superior performance in Chinese-language and modality-biased scenarios, as well as more diverse, concise, and accurate responses.

\textbf{Vertical domain question answering.} While maintaining high competitiveness in general capabilities, MindGPT-4ov also expands its knowledge in vertical domains. In practical applications, it effectively reduces hallucination problems caused by insufficient knowledge yet forced reasoning, while preserving strong cross-domain generalization capabilities. As shown in Fig.~\ref{fig:experiment1}, when addressing questions involving new vertical domain knowledge, MindGPT-4ov provides accurate responses, whereas the baseline model Qwen3-VL lacks the ability to answer such questions accurately.

\textbf{Chinese question answering.} MindGPT-4ov also demonstrates superior performance in Chinese tasks. As shown in Fig.~\ref{fig:exp-benchmark} (a), MindGPT-4ov outperforms others in both general Chinese question answering tasks and Chinese OCR document comprehension. Furthermore, the examples shown in Fig.~\ref{fig:experiment2} indicate that MindGPT-4ov provides precise and focused reasoning in its responses, whereas Qwen3-VL generates redundant and logically flawed answers. This further confirms MindGPT-4ov’s superior capability in Chinese question answering.

\textbf{Modality-biased question answering.} We further demonstrate the MindGPT-4ov's capabilities in cross-modal alignment, noise resistance, and factual output across three typical modality-biased question-answering scenarios: text-image inconsistency, context inconsistency, and missing images. By comparing the responses of MindGPT-4ov and Qwen3-VL on the same test samples, we evaluate the reliability and accuracy of the models in such challenging settings. 

We further illustrate the proposed MindGPT-4ov’s robustness to interference in three typical modality-biased question-answer scenarios: image–text inconsistency, contextual inconsistency, and missing images. By comparing the responses of MindGPT-4ov and Qwen3-VL on the same case, we evaluate the reliability and accuracy of MindGPT-4ov’ outputs in modality-biased question-answer settings.
As illustrated in Fig.~\ref{fig:experiment3}, in Sample-1, which involves cross-modal inconsistency, MindGPT-4ov identifies the visual irrelevance, ignores the erroneous input image, and accurately answers the question.
In Sample-2, which tests context irrelevance, MindGPT-4ov remains unaffected by unrelated interleaved context and correctly responds to both rounds of questions.
In Sample-3 and Sample-4, which evaluate image-missing scenarios, MindGPT-4ov demonstrates stronger factual integrity by explicitly rejecting false outputs when no image is provided.

\textbf{Diverse responses.} We utilize NoveltyBench \cite{noveltybench} to measure the diversity of the models by counting the number of semantically distinct results generated in response to the same prompt, directly reflecting semantic diversity. As shown in Fig.~\ref{fig:exp-benchmark} (b), MindGPT-4ov demonstrates improved performance on NoveltyBench. Furthermore, we present several cases to illustrate the greater diversity in MindGPT-4ov's responses. As depicted in Fig.~\ref{fig:experiment4}, Qwen3-VL exhibits monotonicity in its replies, repeatedly using the ``One of the Harry Potter books is..." with slight redundancy and repetition in expression, while lacking informational expansion. In contrast, MindGPT-4ov's responses avoid repetitive sentence structures and cover more dimensions of information.

\textbf{Concise and accurate responses.} 
An excellent MLLM must not only possess strong general capabilities but also can ensure that its interaction style aligns with human habits. In practical application evaluations, MindGPT-4ov demonstrates outstanding performance in both response accuracy and conciseness, delivering a more human-like interaction experience. When presented with the same input image and question, MindGPT-4ov can directly address the core of the query, providing precise answers while effectively avoiding redundant details. In contrast, the baseline model Qwen3-VL often produces lengthy and irrelevant responses.

We showcase the advantages of MindGPT-4ov's user experience across multiple dimensions. For instance, as illustrated in Fig.~\ref{fig:experiment5}, we collect examples from the general question-answering benchmark MMStar, the mathematical capability benchmark MathVision, and the STEM benchmark MMMU. These examples reveal that Qwen3-VL tends to generate excessively long responses with unnecessary background information, whereas MindGPT-4ov directly identifies the key points, offering concise and accurate descriptions along with correct answers. This human-like response characteristic significantly enhances interaction efficiency and user experience in real-world applications.

Furthermore, to objectively evaluate the conciseness and accuracy of MindGPT-4ov's responses, we compared the response length distributions of MindGPT-4ov-30B-A3B and Qwen3-VL-30B-A3B on MathVista. The results shown in Fig.~\ref{fig:exp-benchmark} (c) indicate that MindGPT-4ov generates shorter responses while achieving an accuracy of 83.3 on MathVista, surpassing Qwen3-VL by 3.2\%.
\section{Conclusion}
This paper proposes MindGPT-4ov, a systematic and end-to-end engineering framework for multimodal large language models tailored to vertical applications. The framework encompasses the entire pipeline of data management, model training, and deployment, aiming to effectively transfer the general capabilities of large multimodal models to vertical domains while mitigating catastrophic forgetting and enhancing user experience. By integrating information density-based data synthesis methods, collaborative curriculum supervised fine-tuning strategies, multi-stage hybrid reward reinforcement learning mechanisms, and efficient training and inference optimization techniques, MindGPT-4ov demonstrates significant advantages in both general and vertical tasks; its performance matches or even surpasses existing advanced multimodal large language models, while also exhibiting excellent user experience and system scalability. MindGPT-4ov is a reproducible, scalable approach for vertical-domain customization that requires no changes to the underlying architecture of pretrained general multimodal large language models, which enables seamless transfer from general capabilities to vertical applications. 
\section{Contributions}

All contributors of this work are listed in alphabetical order by their last names.

\textbf{Core Contributors:}
Chaoqun Du, Feng Gu, Wei He, Qizhen Li, Zide Liu, Xuhao Pan, Chang Ren, Xudong Rao, Chenfeng Wang, Chengjun Yu, Yufei Zheng, Chunpeng Zhou, Xuhan Zhu

\textbf{Contributors:}
Jiawei Chen, Xingru Chen, Handong Cui, Li Gong, Maokui He, Yi Hu, Tianyu Lan, Hengtao Li, Shanshan Li, Xiaobo Liu, Hongli Luo, Jing Luo, Hao Ma, Ning Mao, Lifu Mu, Zhiheng Qu, Tong Sun, Xintian Shen, Haoyi Sun, Qian Wang, Wei Wang, Xin Wang, Lian Wen, Zhichao Wang, Shuo Xie, Hongfu Yang, Jiajun Yang, Shengyu Yao, Mi Yuan, Jiqing Zhan, Hongyuan Zhang, Qi Zhang, Lihao Zheng, Zheng Zhou, Jiaxu Zhu, Yun Zhu

\textbf{Project Leaders:}
Tao Wei,
Pengfei Yu

\textbf{Supervisors:}
Wei Chen,
Pan Zhou

\bibliographystyle{plainnat}
\bibliography{llava}

\end{document}